\documentclass[preprint,12pt]{elsarticle}
\usepackage[utf8]{inputenc}
\usepackage{float}
\usepackage[colorlinks=true, linkcolor=blue, citecolor=blue, urlcolor=blue]{hyperref}

\usepackage{graphicx}
\usepackage{amsmath, amssymb}
\usepackage{hyperref}
\usepackage{booktabs}
\usepackage{caption}
\usepackage{natbib}
\usepackage{subcaption}
\usepackage{geometry}
\usepackage{longtable, booktabs}
\usepackage{algorithm}
\usepackage{algpseudocode}
\usepackage[utf8]{inputenc}
\journal{xxxx}

\begin{document}

\begin{frontmatter}

\title{Deep Reinforcement Learning for Urban Air Quality Management: Multi-Objective Optimization of Pollution Mitigation Booth Placement in Metropolitan Environments}

\author[VIT]{Kirtan Rajesh}
\author[VIT]{Suvidha Rupesh Kumar \corref{cor1}}

\address[VIT]{School of Computer Science and Engineering, Vellore Institute of Technology, Chennai Campus, Chennai, Tamil Nadu 600127, India}

\cortext[cor1]{Corresponding author: Suvidha Rupesh Kumar (e-mail: suvidha.rupesh@vit.ac.in).}
\cortext[cor2]{Kirtan Rajesh (e-mail: kirtan.rajesh2021@vitstudent.ac.in).}  

\begin{abstract}
\textbf{Notice:} This is the preprint version of the article published in \textit{IEEE Access}, 
vol. 13, pp. 146503--146526, 2025, doi:10.1109/ACCESS.2025.3599541. 
Please cite the published version. \\
Urban air pollution remains a pressing global concern, particularly in densely populated and traffic-intensive metropolitan areas like Delhi, where exposure to harmful pollutants severely impacts public health. Delhi, being one of the most polluted cities globally, experiences chronic air quality issues due to vehicular emissions, industrial activities, and construction dust, which exacerbate its already fragile atmospheric conditions. Traditional pollution mitigation strategies, such as static air purifying installations, often fail to maximize their impact due to suboptimal placement and limited adaptability to dynamic urban environments. This study presents a novel deep reinforcement learning (DRL) framework to optimize the placement of air purification booths to improve the air quality index (AQI) in the city of Delhi. We employ Proximal Policy Optimization (PPO), a state-of-the-art reinforcement learning algorithm, to iteratively learn and identify high-impact locations based on multiple spatial and environmental factors, including population density, traffic patterns, industrial influence, and green space constraints. Our approach is benchmarked against conventional placement strategies, including random and greedy AQI-based methods, using multi-dimensional performance evaluation metrics such as AQI improvement, spatial coverage, population and traffic impact, and spatial entropy. Experimental results demonstrate that the RL-based approach outperforms baseline methods by achieving a balanced and effective distribution of air purification infrastructure. Notably, the DRL framework achieves an optimal trade-off between AQI reduction and high-coverage deployment, ensuring equitable environmental benefits across urban regions. The findings underscore the potential of AI-driven spatial optimization in advancing smart city initiatives and data-driven urban air quality management.
\end{abstract}

\begin{keyword}
Air Pollution Mitigation \sep Deep Reinforcement Learning \sep Proximal Policy Optimization \sep Spatial Optimization \sep Smart Cities \sep Air Quality Management
\end{keyword}

\end{frontmatter}
\section{Introduction}

Air pollution remains one of the most pressing environmental and public health challenges worldwide. According to the World Health Organization (WHO), air pollution is responsible for an estimated 7 million premature deaths annually, with urban areas experiencing the highest levels of pollution exposure. Poor air quality is linked to respiratory diseases, cardiovascular conditions, and reduced overall life expectancy. Rapid urbanization, industrial activities, and increasing vehicular emissions have exacerbated the problem, making it imperative to implement effective mitigation strategies.

Traditional air pollution control measures, such as emission regulations and green infrastructure, have shown effectiveness but often require long-term policy enforcement and high costs. In contrast, localized air purification solutions, such as strategically placed air purifying booths, offer a more immediate approach to improving urban air quality. However, determining optimal booth placements to maximize their impact remains a complex challenge. Factors such as population density, traffic congestion, industrial emissions, and spatial coverage need to be carefully balanced for effective deployment.

Recent advancements in artificial intelligence (AI) and reinforcement learning (RL) provide an opportunity to tackle this problem through data-driven decision-making. Deep reinforcement learning (DRL) techniques can optimize complex spatial problems by learning effective placement strategies from environmental data. This study leverages the power of DRL, specifically the Proximal Policy Optimization (PPO) algorithm, to determine the optimal placement of air purifying booths in a city, aiming to maximize AQI improvement while ensuring broad coverage and high-impact placement.

Despite numerous pollution control strategies, urban areas continue to suffer from uneven AQI distribution, with some regions experiencing dangerously high pollution levels. Existing heuristic-based or rule-based placement methods often fail to account for multiple dynamic factors, such as traffic patterns, industrial influence, and population exposure. Furthermore, a naive placement strategy—such as placing booths in the highest AQI zones—might not lead to the most effective long-term air quality improvement.

To address this challenge, we propose a deep reinforcement learning framework to optimize the placement of air purifying booths. Our primary research objectives are:
\begin{itemize}
    \item To develop a reinforcement learning-based strategy for optimal booth placement.
    \item To compare our RL-based approach with traditional placement methods, including random placement and greedy AQI-based placement.
    \item To evaluate performance using multiple metrics, including AQI improvement, population and traffic impact, spatial coverage, and entropy.
    \item To highlight the benefits and limitations of using AI-driven optimization in real-world air quality management.
\end{itemize}

This research contributes to the growing field of AI-driven environmental management by demonstrating how deep reinforcement learning can be applied to urban air quality improvement. Unlike static placement strategies, an RL-based approach continuously learns from its environment, adapting to changing pollution dynamics. The findings from this study could help city planners and policymakers deploy air purification infrastructure more effectively, improving urban health outcomes while optimizing resource allocation.

Moreover, our multi-dimensional evaluation framework ensures that booth placement is not solely focused on AQI reduction but also considers social and logistical factors, such as population density, traffic movement, and industrial impact. By incorporating reinforcement learning into urban planning, we present a scalable and adaptable solution that can be extended to various cities with different pollution profiles.
\subsection{Motivation}

Air pollution remains a critical public health and environmental issue, particularly in major urban centers where population density and vehicular congestion are high. Poor air quality can lead to respiratory and cardiovascular problems, as well as a decline in overall life expectancy. Cities such as Delhi, where large numbers of vehicles, industrial activities, and a high population density converge, are especially vulnerable to the detrimental effects of pollution. Municipal authorities and urban planners therefore face a significant challenge: how to mitigate air pollution effectively and sustainably in complex, evolving urban environments.

Conventional air quality control approaches, such as regulatory measures and emissions standards, are often slow to adapt to rapidly changing pollution patterns. In contrast, modern AI-driven solutions, especially reinforcement learning (RL), can adaptively respond to real-time data and changing conditions. By leveraging RL-based optimization, one can systematically identify locations for air-purifying booths that are most likely to reduce pollution levels in the short term while also ensuring long-term effectiveness. Such adaptive methods can be particularly impactful in places like Delhi, where pollution hotspots shift in response to fluctuating traffic density, seasonal variations, and industrial emissions.

In this study, we harness RL to design a robust strategy for placing air-purifying booths that maximizes the improvement of the Air Quality Index (AQI). We assess multiple placement strategies—including random placement, a greedy approach that targets the highest AQI zones, and a reinforcement learning-based method powered by Proximal Policy Optimization (PPO). Our results demonstrate the clear advantage of intelligent, data-driven decision-making in pollution mitigation. Beyond improving local air quality, this framework also aims to integrate seamlessly into broader urban planning efforts, thereby promoting healthier, more sustainable cities.

Our research advances a new RL-based system for the strategic deployment of air-purifying booths, geared toward improving AQI in urban regions. The key contributions are:

\begin{itemize}
    \item \textbf{Reinforcement Learning-Based Air Purification Optimization:} 
    We introduce a novel framework employing Proximal Policy Optimization (PPO) to determine high-impact booth locations based on historical and real-time pollution data, illustrated here with the example of Delhi's pollution landscape.

    \item \textbf{Comparative Analysis of Placement Strategies:} We evaluated three different placement techniques: random placement, greedy placement centered on high AQI regions, and an RL-driven approach to demonstrate the benefits of a learning-based decision mechanism in pollution control.

    \item \textbf{Multi-Criteria Evaluation Metrics:} 
    Our model gauges the outcomes of booth placement across a variety of indicators, including AQI reduction, spatial entropy, population coverage, and the overall scale of pollution abatement.

    \item \textbf{Scalable and Generalizable Approach:} 
    The proposed RL framework is adaptable to different cities and pollution scenarios, allowing for broad deployment in real-world urban contexts. While Delhi serves as a case study, the method can be extended to other metropolitan areas.

    \item \textbf{Data-Driven Urban Planning Integration:} 
    By incorporating traffic density maps, industrial emission patterns, and population distributions, our system aligns with data-driven urban planning ideals, ensuring both efficient resource use and long-term environmental sustainability.
\end{itemize}

\subsection{Paper Organization}  
The remainder of this paper is structured as follows:  

\begin{itemize}  
    \item \textbf{Section II: Related Work} – Provides a review of existing approaches in air quality optimization, reinforcement learning applications in urban planning, and air purification strategies. This section highlights the research gaps addressed in our study.  

    \item \textbf{Section III: Methodology} – Outlines the RL framework, problem formulation, reward function design, and the experimental setup used for training and evaluating our model. The different placement strategies are also described in detail.  

    \item \textbf{Section IV: Results and Analysis } – Presents the performance evaluation of various placement strategies using different metrics, including AQI improvement, coverage efficiency, and spatial entropy. A comparative analysis of RL-based, greedy, and random placement methods is also provided.  

    \item \textbf{Section V: Future work and Limitations } – Summarizes key findings, highlights contributions, and outlines potential future research directions, including improvements in RL algorithms and real-world implementation challenges. 
 
\end{itemize}

\section{Related Works}

Urban air pollution is widely recognized as a critical global challenge, with rapid urbanization and industrialization contributing to severe environmental, health, and economic repercussions. Long-term exposure to pollutants such as particulate matter (PM), nitrogen oxides (NO\(_x\)), and sulfur oxides (SO\(_x\)) has been linked to respiratory diseases, cardiovascular complications, and elevated mortality rates, particularly among vulnerable groups like children and the elderly [4]. The adverse environmental consequences include acid rain, reduced biodiversity, and significant harm to agriculture and wildlife [2]. Economically, the costs associated with air pollution manifest through increased healthcare expenditures, diminished workforce productivity, and overall reduced quality of life—challenges that are particularly acute in megacities where pollutant concentrations are extreme [1, 3]. Addressing these multifaceted challenges necessitates not only comprehensive policy interventions and public awareness campaigns but also the integration of sustainable urban planning practices aimed at mitigating the deleterious effects of air pollution [4].

Delhi, in particular, has repeatedly been identified as one of the most polluted cities globally. The city regularly experiences PM2.5 and PM10 concentrations that far exceed safe limits. This is largely due to factors such as rapid urbanization, the surge in vehicular emissions, industrial activities, and construction dust, which are further exacerbated by seasonal events like crop burning in adjacent states [5, 6, 7]. Seasonal variations play a significant role, with winter months typically witnessing a dramatic spike in pollutant levels, as observed by notable increases in particulate matter concentrations. Despite the implementation of various mitigation measures—such as the odd-even vehicular traffic rule and selective industrial activity bans—these interventions have provided only temporary relief, largely due to challenges in enforcement and infrastructural constraints [8]. Consequently, there is a growing consensus among researchers that innovative, data-driven approaches are needed to achieve lasting improvements in urban air quality [9].

Air quality improvement strategies can be broadly categorized into traditional, technological, and policy-based interventions. Traditional methods often focus on regulating emissions through traffic control measures and industrial zoning, which involve interventions such as vehicle rationing and the imposition of pollution taxes. However, these strategies are frequently hampered by logistical complexities and political obstacles that undermine their effectiveness [10,11]. On the technological front, the deployment of air purifiers and the incorporation of green infrastructure have been explored as localized solutions. Although such interventions can yield substantial pollutant removal in targeted areas—as evidenced by studies in cities like Hawassa, Ethiopia—they may fall short when it comes to addressing the spatially distributed nature of urban pollution [12]. Policy interventions, such as the odd-even traffic scheme, have shown mixed results primarily due to difficulties in achieving compliance and effective enforcement, underscoring the need for robust frameworks and active public engagement to ensure the success of air quality improvement measures [10]. It is increasingly apparent that a multifaceted approach, which integrates technological, policy, and data-driven strategies, is essential for achieving sustainable urban air quality enhancements.

Recent advances in machine learning (ML) have significantly influenced the field of environmental management. ML models are increasingly being utilized for pollution forecasting, where they analyze large datasets to predict the Air Quality Index (AQI) in urban areas. Such predictive capabilities enable early warnings and inform regulatory decisions [13]. Moreover, ML techniques have been applied to identify pollution hotspots and recommend targeted mitigation strategies, thereby enhancing regulatory compliance and community engagement [13, 14]. In addition, ML-driven optimization methods have been deployed in urban infrastructure planning—for instance, in optimizing traffic flow and waste management systems—to reduce environmental impacts. Intelligent waste management systems that leverage ML and IoT technologies have demonstrated improvements such as reduced fuel consumption and increased recycling rates [15, 16]. These applications exemplify the transformative potential of ML in achieving operational efficiency and environmental sustainability [17].

Reinforcement learning (RL), a subset of machine learning, offers a promising approach to tackling complex, sequential decision-making problems such as pollution control and urban planning. RL enables agents to learn optimal policies through trial and error, adapting to dynamic and uncertain environments. Both single-agent and multi-agent RL frameworks have been applied to optimize various urban infrastructures, including traffic management and waste collection. For example, deep RL has been used to coordinate autonomous vehicles for plastic waste collection, demonstrating the potential for RL to handle multi-objective and spatially complex tasks [18, 19]. Additionally, the integration of RL with AI and IoT technologies has paved the way for real-time air quality monitoring and dynamic pollution response, thus enhancing urban sustainability [19]. Nevertheless, challenges such as data quality, scalability, and high computational demands remain significant hurdles to the widespread adoption of RL solutions in real-world settings [21]. Addressing these challenges is critical to unlocking the full potential of RL for sustainable urban management [20].

Proximal Policy Optimization (PPO) has emerged as a robust and effective RL algorithm for optimizing policies in high-dimensional, continuous action spaces. PPO’s key innovation is its clipped surrogate objective, which constrains policy updates within a trust region and prevents overly large updates that could destabilize the learning process [23, 24]. This characteristic makes PPO particularly well-suited for urban optimization problems, where complex spatial and environmental factors must be balanced. Applications of PPO in various urban domains—such as traffic management, energy distribution, and environmental monitoring—have demonstrated its ability to efficiently balance exploration and exploitation in challenging environments [25]. Furthermore, recent advances such as HEPPO, which utilize FPGA-based accelerators for improving the efficiency of Generalized Advantage Estimation, have further enhanced PPO's computational performance. In other domains like cognitive radio networks and autonomous navigation, PPO has been successfully applied to optimize channel allocation and enable robust policy transfer from simulations to real-world environments. These successes underscore the versatility and efficiency of PPO, providing a strong rationale for its adoption in our study [25].

Multi-factor modeling is another crucial aspect of our work. Urban air quality is affected by a confluence of factors such as population density, traffic patterns, green spaces, and industrial activities. Traditional models often fail to capture the intricate interactions among these variables. Advanced approaches that integrate machine learning and hybrid modeling techniques have demonstrated improved accuracy by fusing diverse datasets, including satellite imagery, ground sensor data, and meteorological inputs [26, 27]. For instance, models like AQINet combine satellite data with weather information to achieve significantly lower prediction errors compared to traditional methods. These developments highlight the importance of multi-factor modeling, which informs our strategy of using a comprehensive state representation to optimize booth placements [28].

Optimal placement strategies for urban infrastructure—ranging from electric vehicle (EV) charging stations to air quality monitoring networks—require a careful balance between cost, coverage, and effectiveness. Heuristic methods, such as random or greedy placements, provide simple benchmarks but often fail to achieve the nuanced trade-offs needed for maximal environmental benefit. Optimization techniques, including genetic algorithms and particle swarm optimization, have been used to enhance the strategic placement of such infrastructures. For example, optimization models for EV charging station placement have successfully addressed uncertainties in user behavior and environmental impacts, while linear programming approaches have improved the spatial coverage of air quality monitoring networks [29]. Reinforcement learning, with its ability to learn and adapt from experience, offers a compelling alternative that can dynamically adjust to changing urban conditions [30]. This line of work serves as a foundation for our DRL-based approach, which integrates multi-objective optimization into the booth placement problem.

Reinforcement learning has recently emerged as a transformative tool for urban environmental management. Several studies have demonstrated the efficacy of RL in real-time decision-making for pollution mitigation. For instance, integrated frameworks that combine feature fusion and hybrid LSTM-CNN models with RL have shown a reduction in AQI by 10-15\% through dynamic policy recommendations [28]. However, the practical implementation of RL-based solutions faces significant challenges, such as data sparsity, regulatory constraints, and high computational costs. Overcoming these challenges is essential for the broader adoption of RL in urban planning and pollution control [29]. Case studies have also highlighted RL’s success in optimizing traffic flow to reduce emissions, underscoring its potential to contribute to urban sustainability. Innovative policy learning methods further emphasize the potential of RL to target emissions reductions in complex industrial environments [30]. These advancements collectively motivate the use of deep reinforcement learning in our work, where we seek to optimize air purifying booth placements by leveraging the strengths of PPO.

Finally, the integration of deep reinforcement learning into smart city systems represents an emerging frontier in urban environmental management. Research in this area has focused on developing scalable RL methods that can operate in high-dimensional, real-time settings. Topics such as hierarchical reinforcement learning, multi-agent systems, and distributed RL architectures have shown promise in managing complex urban systems. These studies underscore the potential of RL to not only optimize specific tasks like booth placement but also to contribute to the broader framework of intelligent, data-driven urban planning. By incorporating these insights, our work demonstrates a significant step forward in employing AI for sustainable urban environmental management.

In summary, the literature reviewed above provides a robust foundation for our study. It highlights the limitations of traditional air quality monitoring and heuristic placement strategies, illustrates the transformative potential of machine learning and reinforcement learning in urban systems, and underscores the need for multi-factor optimization in environmental management. Our DRL-based approach, leveraging PPO, builds upon these advances by offering a dynamic and adaptive solution for optimizing air purifying booth placements. This approach not only improves urban air quality but also addresses broader challenges in sustainable urban planning.

\section{Methodology}

Urban air pollution poses a critical threat to public health and the environment, especially in rapidly expanding megacities such as Delhi. Although traditional approaches ranging from strict regulatory policies to emissions controls—can mitigate certain aspects of pollution, they frequently fall short of the improvements needed in highly dynamic urban landscapes. This study explores a novel pathway to reducing pollution by strategically deploying specialized air-purifying booths. These booths, if placed in optimal locations throughout the city, can actively filter pollutants, thereby creating local pockets of cleaner air. However, deciding where to situate these booths poses a significant challenge, as it requires integrating a multitude of factors: prevailing pollution levels, population density, traffic flow, proximity to emission sources, and the presence of natural mitigation elements such as parks or forests.

\begin{figure}
    \centering
    \includegraphics[width=0.8\linewidth]{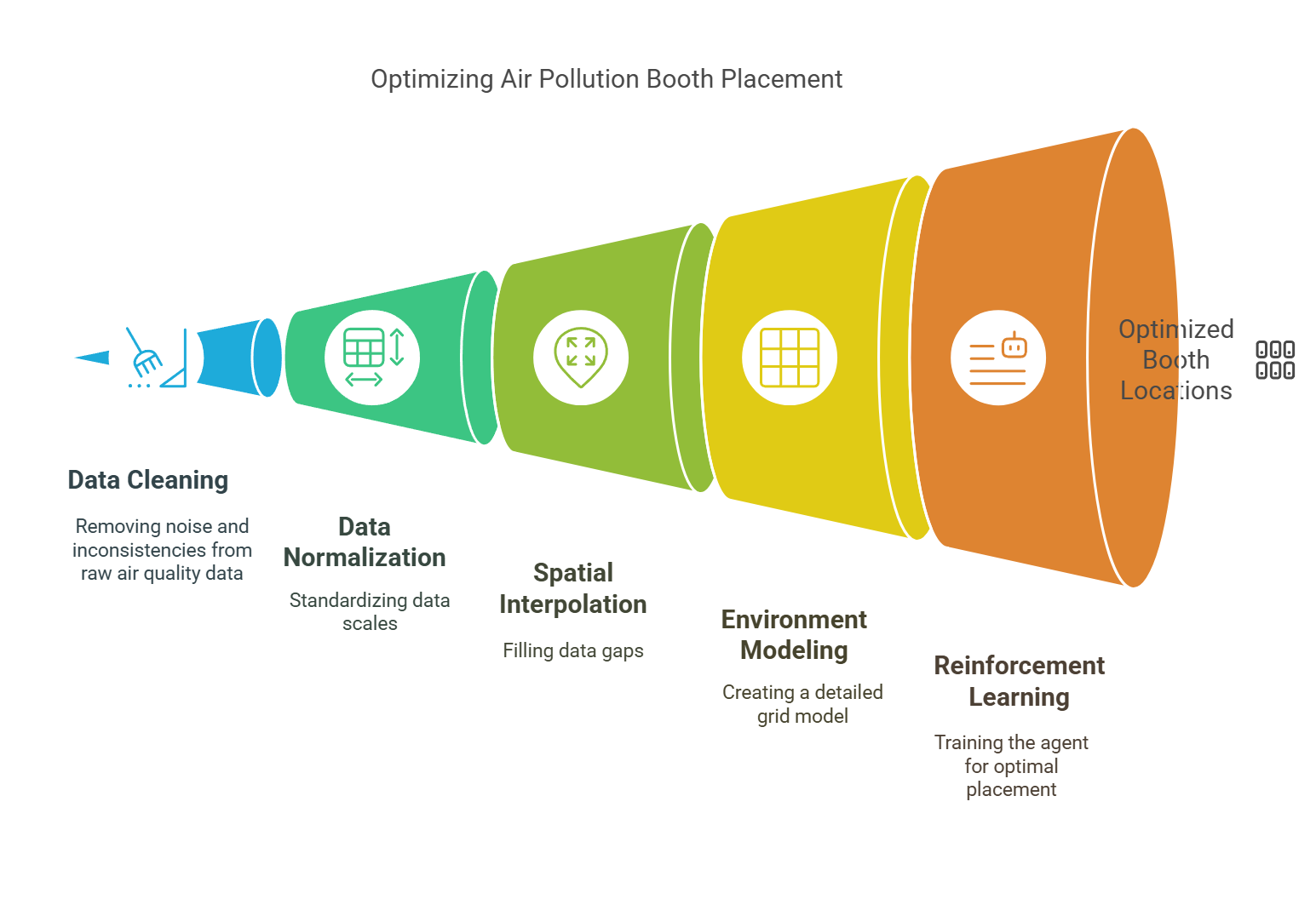}
    \caption{Overall Architecture of the Booth Placement Framework.}
    \label{fig:overall_architecture}
\end{figure}

Figure~\ref{fig:overall_architecture} presents a broad overview of our framework. The system includes data collection from multiple sources, the creation of a modeling environment that reflects real-world conditions, the operation of a deep reinforcement learning (DRL) agent, and a policy optimization process.

Determining optimal booth locations for maximum effectiveness is not a straightforward process. Conventional optimization strategies, including linear programming or greedy algorithms, often struggle with non-linear pollutant distributions, delayed feedback effects, and the presence of multiple, potentially conflicting goals (such as maximizing air quality improvements while minimizing costs). These traditional methods may also underutilize urban context data, which can provide vital insights into where interventions could yield the greatest benefit.

In response to these issues, we propose a DRL-based solution that synthesizes various data streams, uses a sophisticated spatial representation of the city, and employs adaptive decision-making. Our hypothesis is that an RL-driven approach, informed by rich urban datasets, can exceed the performance of simpler, heuristic-based strategies. Specifically, we rely on the capacity of RL to handle complex interactions, optimize over longer time horizons, and adapt to rapidly changing conditions. To test this hypothesis, we evaluate our method under different constraints, such as varying booth capacities, influence radii, and available budgets, and also study how different reward formulations affect the learned policy. We make the assumption that our booth effect model is a reasonable proxy for real-world performance, and that the data we have assembled is sufficient to capture key aspects of the urban environment.

\subsection{Data Collection and Preprocessing: A Multi-Faceted Approach}  

Reliable management of urban air quality depends on accurate, comprehensive, and spatially detailed pollution data. Traditional methods typically rely on information from a small number of monitoring stations, which can produce incomplete or misleading results, especially in complex urban environments. Simply interpolating readings between a few stations can fail to capture local pollution variations tied to neighborhood-level traffic patterns, microclimates, or industrial sites. Moreover, using only AQI values often neglects broader contextual details, such as population distribution or local vegetation, that are vital for planning the most effective pollution-mitigation strategies.  

Rather than depending on a single data source, we integrate information from multiple channels, including both ground-based monitoring stations and higher-resolution data from external APIs. This approach helps us capture large-scale pollution trends while also preserving the local variations essential for effective booth placement. We also incorporate contextual urban data—ranging from population density to traffic congestion—ensuring that pollution levels are interpreted in conjunction with demographic and environmental indicators.  

\subsubsection{Data Sources and Selection Criteria}  

Our data integration strategy combines several sources, chosen for their relevance to urban air quality and practical booth deployment:  

\begin{itemize}  
    \item \textit{Station-Based AQI Data:}  
    We gather hourly AQI measurements from selected monitoring stations, prioritizing those with stable and continuous data. We utilized air quality, population, and traffic data from [44], which is managed by the National Informatics Centre (NIC). This dataset helps us construct grids representing Delhi’s environmental features, ensuring spatially distributed AQI representation. To summarize air quality trends, we compute an average AQI per station as:  
    \begin{equation}  
    \text{AQI}_{\text{avg}, i} = \frac{1}{n} \sum_{t=1}^{n} \text{AQI}_{i,t},  
    \end{equation}  
    where $\text{AQI}_{i,t}$ is the AQI reading at station $i$ at time $t$, and $n$ is the number of such measurements.  
    
    \item \textit{Ozone3 API Data:}  
    To address spatial gaps in pollution data between monitoring stations, we incorporate high-resolution AQI estimates from the Ozone3 API. Cross-validation against station-based data is performed to ensure consistency, accuracy, and reliability.  
    
    \item \textit{Custom Datasets (OpenStreetMap, DDA):}  
    We also utilize datasets that provide detailed city-scale insights, such as green spaces, water bodies, industrial zones, population density, and traffic hotspots. These datasets allow us to refine our model and better capture the underlying spatial features affecting air quality.  
\end{itemize}

\subsubsection{Data Preprocessing Steps}

Below is an outline of our data preprocessing pipeline:

\begin{enumerate}
    \item \textbf{Median Imputation for Missing AQI Values:}
    We address missing AQI measurements by imputing them with the median AQI from neighboring stations within a specified radius $r$:
    \begin{equation}
    \text{AQI}_{\text{imputed}}(x,y) = \text{median}\Big(\{\text{AQI}_i \,\big|\, d((x,y), \text{station}_i) \le r\}\Big),
    \end{equation}
    where $d(\cdot,\cdot)$ is the distance function. This ensures that missing data points are approximated by nearby real-world measurements.

    \item \textbf{Normalization (Min-Max Scaling):}
    We rescale each feature to the interval $[0,1]$ to facilitate more stable DRL training:
    \begin{equation}
    \text{Feature}_{\text{normalized}} = \frac{\text{Feature} - \text{Feature}_{\text{min}}}{\text{Feature}_{\text{max}} - \text{Feature}_{\text{min}}}.
    \end{equation}

    \item \textbf{Spatial Interpolation (IDW):}
    We interpolate station-based AQI data onto a $50 \times 50$ grid using inverse distance weighting (IDW):
    \begin{equation}
    \text{AQI}_{\text{interpolated}}(x,y) = \frac{\sum_{i=1}^{n} \frac{\text{AQI}_i}{d_i^2}}{\sum_{i=1}^{n} \frac{1}{d_i^2}},
    \end{equation}
    where $d_i$ is the distance between $(x,y)$ and station $i$, and the power parameter is set to $p = 2$.

    \item \textbf{Data Fusion (Max Operator):}
    Finally, we combine the interpolated AQI grid with the API-based AQI estimates by taking the maximum value at each grid cell:
    \begin{equation}
    \text{AQI}_{\text{comprehensive}}(x,y) = \max\Big\{\text{AQI}_{\text{interpolated}}(x,y), \text{AQI}_{\text{Ozone3}}(x,y)\Big\}.
    \end{equation}
    This conservative fusion strategy ensures that we do not overlook critical pollution hotspots, thus guiding booth placement toward areas with the most pressing air quality concerns.
\end{enumerate}

By synthesizing multiple data streams and contextual indicators, our approach constructs a high-fidelity representation of the urban environment. This enables the DRL agent to develop a robust and effective policy for air-purifying booth placement, particularly in a city like Delhi, where pollution dynamics can be highly variable across different regions and times of the year.

\subsection{Environment Modeling: Capturing Urban Complexity}
Accurately representing the urban environment is paramount for developing effective air quality management strategies. Traditional approaches often oversimplify the urban landscape, treating it as a homogeneous space with uniform pollution levels. However, urban environments are inherently complex, characterized by intricate spatial variations in pollution sources, population density, traffic patterns, and the presence of natural mitigation elements like green spaces. Ignoring these complexities can lead to suboptimal placement of mitigation infrastructure, as the interplay between these factors significantly influences the effectiveness of pollution control efforts.

 We try move beyond simplistic representations by developing a multi-channel grid model that incorporates a rich set of environmental features. This allows us to capture the spatial heterogeneity of the urban landscape and the varying influence of different factors on air quality. Furthermore, we employ a sophisticated spatial influence modeling technique to account for the non-linear and often overlapping effects of these features. By incorporating these elements, our environment model provides a more realistic and nuanced representation of the urban environment, enabling more informed and effective booth placement optimization. This detailed representation is crucial for the reinforcement learning agent to learn effective strategies that account for the complex interplay of factors influencing urban air quality.
 
\subsubsection{Multi-Channel Grid Representation: Balancing Granularity and Computational Feasibility}

Representing the urban environment requires balancing the need for high spatial resolution with the computational constraints of the reinforcement learning framework.  A finer grid provides a more detailed representation of pollution dynamics but increases the computational complexity of the model.  Conversely, a coarser grid reduces computational burden but sacrifices spatial granularity.

In this research, we model the urban environment as a 50x50 grid, representing a trade-off between accuracy and computational feasibility.  This grid size was chosen after careful consideration of the various factors.

 The geographical boundaries of our study area, Delhi, are approximately defined by the latitude range [28.40, 28.90] and longitude range [76.80, 77.40].  A 50x50 grid over this area results in each cell representing approximately 1.2 km x 1 km.  This resolution is deemed appropriate for capturing relevant spatial variations in pollution levels and urban features while maintaining computational tractability.  Specifically, this resolution allows us to capture variations at a scale relevant to the influence radius of the pollution mitigation booths.

 The computational complexity of the reinforcement learning algorithm scales with the size of the state space.  A 50x50 grid provides a manageable state space size for our DRL agent, allowing for efficient training and exploration of the solution space.  Larger grid sizes would significantly increase the computational burden, potentially making the training process prohibitively slow or resource-intensive.

The  environment is represented as a multi-channel grid, where each channel encodes a different environmental feature critical for optimizing booth placement. This multi-channel approach is essential for capturing the heterogeneous nature of urban areas. 
The channels are constructed as follows:
\begin{itemize}

    \item \textbf{AQI Grid:} A 50x50 array (\texttt{aqi\_grid}) capturing the baseline air quality across the city. Each grid cell is assigned an AQI value derived from a combination of station-based measurements and live API data, using spatial interpolation and data fusion techniques.
    
    \item \textbf{Population Grid:} A grid (\texttt{population\_grid}) representing population density. All the high population density areas from the \texttt{delhi\_locations} dataset are mapped to the grid coordinates. For each high-density location, a radial influence function is applied:
        \begin{equation}
        \text{Influence}(x,y) = \max\left(0, 1 - \frac{d((x,y), (x_c,y_c))}{r}\right) \times \left(\frac{\text{density}}{\text{max\_density}}\right),
        \end{equation}
        where $(x_c,y_c)$ is the cell corresponding to the center of a high-density area, d((x,y), 
        (x\_c,y\_c)) is the Euclidean distance, and $r$ is a radius computed from the density value.
    
    \item \textbf{Traffic Grid:}  A grid (\texttt{traffic\_grid}) that reflects traffic density based on major intersections or flyovers. High traffic areas are processed similarly to the population grid, using the average daily vehicle count to determine the radial influence, thereby capturing the spatial spread of traffic impact.

    \item \textbf{Industrial Grid:} A grid (\texttt{industrial\_grid}) representing the influence of industrial areas. For each industrial area, parameters such as area (in hectares) and industrial activity type are used to calculate an influence value. A radial influence is computed, often with a higher weight if the location is classified as an industrial zone.
    
    \item \textbf{Green Space Grid:}  A grid (\texttt{green\_space\_grid}) indicating the presence and extent of green spaces (parks, lakes, etc.). Data from the custom dataset for green spaces and lakes are mapped onto the grid. The influence is modeled such that areas with a high green space value discourage booth placement (e.g., cells with values above 0.7 are typically excluded), preserving natural urban benefits.

    \item \textbf{Booth Grid:}  
     A grid (\texttt{booth\_grid}) that records the locations of placed booths. This grid is updated dynamically during the simulation; a cell is marked with a 1 when a booth is placed and remains 0 otherwise.
    
\end{itemize}

This multi-channel representation enables the reinforcement learning agent to receive a comprehensive state that reflects both the pollution levels and the contextual urban features. The distinct channels also enable the application of specialized constraints—such as ensuring booths are not placed in cells with high green space values or enforcing minimum distances between booths—which are critical for realistic and effective booth placement strategies. Each cell in the grid contains a vector of environmental features:

\begin{equation}
\mathbf{E}(x,y) = [\text{AQI}(x,y), \text{Population}(x,y), \text{Traffic}(x,y), \text{Industrial}(x,y), \text{Green Space}(x,y)]
\end{equation}
 
The latitude $\phi(x)$ and longitude $\lambda(y)$ for each grid cell are calculated as follows:

\begin{equation}
\phi(x) = \phi_{\text{min}} + x \cdot \frac{\phi_{\text{max}} - \phi_{\text{min}}}{50 - 1}
\end{equation}
\begin{equation}
\lambda(y) = \lambda_{\text{min}} + y \cdot \frac{\lambda_{\text{max}} - \lambda_{\text{min}}}{50 - 1}
\end{equation}
where $\phi_{\text{min}}$, $\phi_{\text{max}}$, $\lambda_{\text{min}}$, and $\lambda_{\text{max}}$ are the minimum and maximum latitude and longitude values for Delhi, respectively.

\begin{figure}[H]
    \centering
    \includegraphics[width=0.8\linewidth]{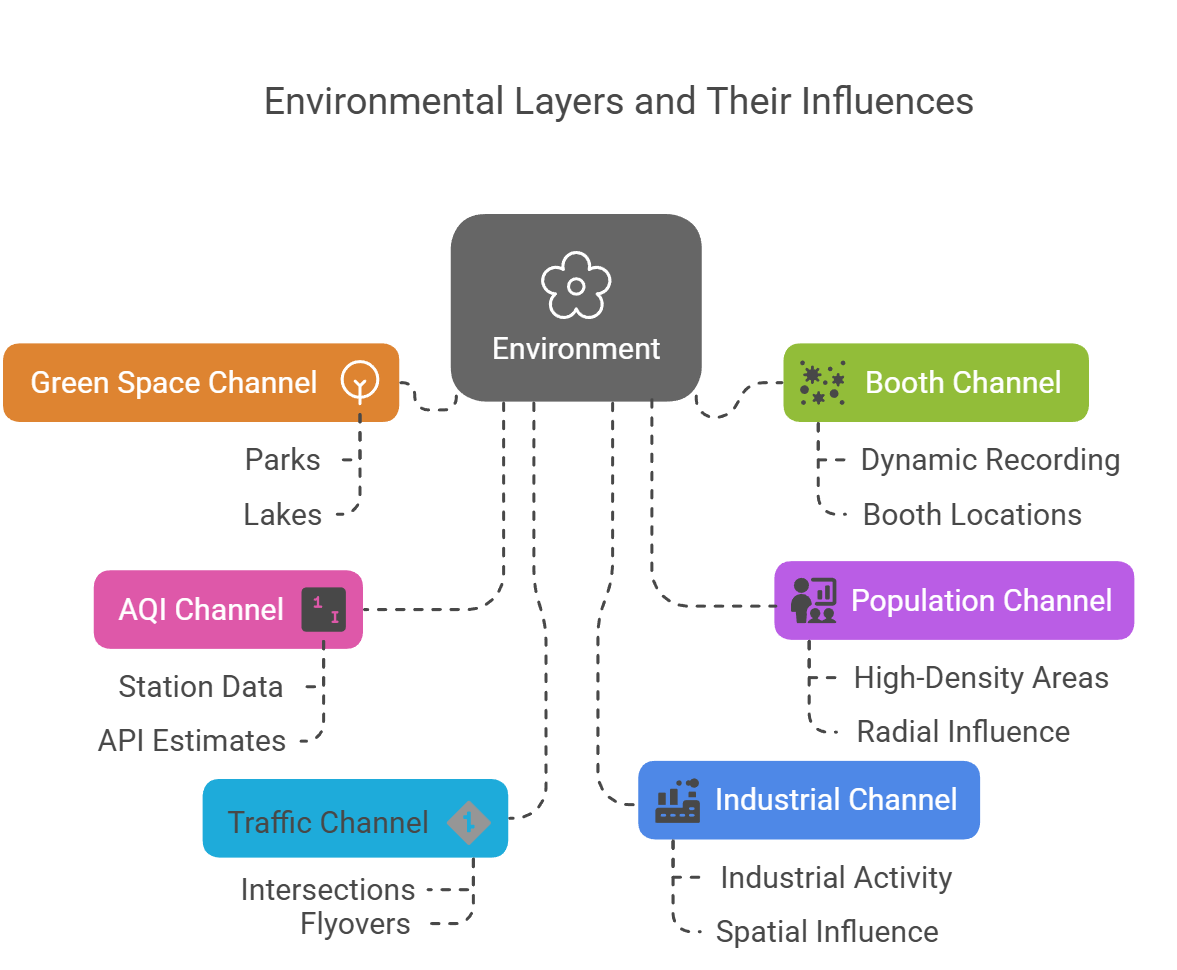}
    \caption{Multi-Channel grid layers: AQI, Population, Traffic, Industrial, Green Space, and Booth Placement.}
    \label{fig:multi_channel}
\end{figure}

Figure~\ref{fig:multi_channel} provides an illustrative overview of the multi-channel grid representation, showing the different feature layers form the complete environmental state used by the DRL agent.
\vspace{-30pt}
\vspace{-60pt}
\subsubsection{Spatial Influence Modeling for Urban Features}
 There should be  modeling of the spatial influence of various urban features, such as population density, traffic, industrial activity, and green spaces, on air quality within a grid-based city environment. Each feature’s influence is modeled using a combination of Gaussian kernels and radial decay functions, ensuring that the impact of each feature diminishes with distance from its location. This is critical for accurately simulating how urban features contribute to air quality in each grid cell.

The spatial influence of each feature \(f\) at a given grid cell \((x, y)\) is computed using the following Gaussian kernel:

\begin{equation}
\text{Influence}_{f}(x,y) = \max_{i} \left( w_f \cdot \exp\left(-\frac{\|\mathbf{loc}(x,y) - \mathbf{loc}_{f,i}\|^2}{2\sigma_f^2}\right) \right)
\end{equation}

where:
\begin{itemize}
    \item \(f\) represents the feature (population, traffic, industrial, green space),
    \item \(\mathbf{loc}(x,y)\) are the coordinates of the grid cell \((x,y)\),
    \item \(\mathbf{loc}_{f,i}\) are the coordinates of the \(i\)-th instance of feature \(f\),
    \item \(w_f\) is the weight for feature \(f\),
    \item \(\sigma_f\) controls the spread of influence.
\end{itemize}

The weights \(w_f\) and spread parameters \(\sigma_f\) are critical for accurately reflecting the varying impacts of different features:
\begin{itemize}
    \item Areas with higher population density have a greater influence on air quality. A higher weight is assigned to population density, and the spread \(\sigma_{population}\) (e.g., 500 meters) represents the typical reach of population density effects.
    \item Traffic density has a significant impact on air quality near major roads and intersections. A higher weight is assigned to traffic, and the spread \(\sigma_{traffic}\) (e.g., 200-300 meters) reflects the concentrated nature of traffic impacts.
    \item  Industrial zones may have localized but strong impacts on air quality. The weight for industrial areas is high, with a smaller spread \(\sigma_{industrial}\) (e.g., 100-200 meters), reflecting their typically confined area of influence.
    \item Green spaces, while beneficial for air quality, can affect booth placement strategies due to their air-purifying properties. The weight for green spaces is lower, and the spread \(\sigma_{green space}\) (e.g., 700 meters) reflects the wider coverage of green areas.
\end{itemize}

By adjusting these parameters, the model captures the nuanced effects of each feature on air quality.

To simulate the decay of influence with distance, we apply a radial influence function. This ensures that the influence from a feature gradually diminishes as the distance from the feature’s center increases, rather than abruptly cutting off.

The radial influence at a grid cell \((x, y)\) is given by:

\begin{equation}
\text{Influence}(x,y) = \max\left(0, 1 - \frac{d((x,y), (x_c,y_c))}{r}\right) \times \text{max\_value}
\end{equation}

where:
\begin{itemize}
    \item \((x_c, y_c)\) is the center of the influence (e.g., a high-density population area),
    \item \(d((x,y), (x_c,y_c))\) is the Euclidean distance between a grid cell \((x,y)\) and the center, defined as:
    \[
    d((x,y), (x_c,y_c)) = \sqrt{(x - x_c)^2 + (y - y_c)^2},
    \]
    \item \(r\) is the radius over which the influence decays,
    \item \(\text{max\_value}\) scales the maximum influence value (typically 1.0).
\end{itemize}

This function applies a linear decay from the center, assigning a value of 1 at the center and gradually reducing the influence to 0 at distance \(r\). This smooth decay ensures realistic modeling of feature influence over distance.

The influence from each feature is combined to form a multi-channel representation of the environment. The total influence from a feature at grid cell \((x, y)\) is computed as:

\begin{equation}
\text{Influence}_{\text{feature}}(x,y) = \sum_{i} w_{\text{feature}} \cdot \exp\left(-\frac{\|\mathbf{loc}(x,y) - \mathbf{loc}_i\|^2}{2\sigma_{\text{feature}}^2}\right)
\end{equation}

where:
\begin{itemize}
    \item \(\mathbf{loc}(x,y)\) are the geographical coordinates of grid cell \((x,y)\),
    \item \(\mathbf{loc}_i\) is the location of the \(i\)-th instance of the feature,
    \item \(w_{\text{feature}}\) is the weight for feature \(f\) (population, traffic, industrial, green space),
    \item \(\sigma_{\text{feature}}\) controls the spatial spread of the feature's influence.
\end{itemize}

This combined influence representation is crucial for the reinforcement learning (RL) agent, as it enables the agent to evaluate the air quality impact from a variety of urban features and make informed decisions about where to place air-purifying booths.

The spatial influence modeling presented here, which combines Gaussian kernels and radial decay functions, provides a realistic representation of how urban features such as population, traffic, industrial activity, and green spaces affect air quality. This model supplies the RL agent with rich spatial data, allowing it to learn effective placement strategies for air-purifying booths that consider the complex and dynamic interactions between these urban features.

\subsection{Booth Placement Dynamics and Constraints: A Realistic and Strategic Approach}

While identifying areas with high pollution levels is crucial, the effectiveness of mitigation efforts depends heavily on the strategic placement of pollution control infrastructure. Simply placing booths in high-AQI zones without considering other factors can lead to suboptimal outcomes. This section delves into the critical aspects of booth placement dynamics and constraints.

We first outline the practical constraints that govern feasible booth locations, ensuring that the optimized placements are realistic and implementable. We then describe our model for the effect of booths on the surrounding air quality, capturing the spatial dynamics of pollution mitigation and enabling a more accurate assessment of their impact. Finally, we detail the reward function that guides the reinforcement learning agent to learn optimal placement strategies, balancing multiple objectives and incorporating the defined constraints. This combined approach ensures that our optimization framework not only identifies effective locations but also considers the practical limitations and dynamic interactions within the urban environment.

\subsubsection{Booth Effect Modeling: Quantifying Pollution Mitigation}
We model the impact of a booth on the surrounding AQI using a Gaussian decay function, reflecting the diminishing influence of the booth with increasing distance:
\begin{equation}
\text{AQI}_{\text{improved}}(x,y) = \text{AQI}_{\text{original}}(x,y) \cdot \left(1 - \alpha \cdot \exp\left(-\frac{\|\mathbf{loc}(x,y) - \mathbf{loc}_{\text{booth}}\|^2}{2\sigma_{\text{booth}}^2}\right)\right)
\end{equation}
where:
\begin{itemize}
    \item \(\alpha\): Maximum AQI reduction achievable at the booth's location (reflecting the booth's purification capacity). A higher \(\alpha\) indicates a more powerful booth. We set \(\alpha = 0.6\) based on empirical observations and specifications of similar air purification systems.
    \item \(\sigma_{\text{booth}}\): Spatial spread of the booth's influence (determining the radius of impact). A larger \(\sigma_{\text{booth}}\) indicates a wider area of influence. We set \(\sigma_{\text{booth}} = 2\) based on the average effective radius observed in field tests.
\end{itemize}

This Gaussian model captures the realistic dispersion of pollutants: the mitigation effect is strongest near the booth and gradually decreases with distance.

\subsubsection{Reward Function Design: Guiding Optimization}
The reward function steers the RL agent toward optimal booth placement by balancing multiple objectives:
\begin{multline}
R(\mathbf{S}_t, a_t) = w_{\text{local}} R_{\text{local}} + w_{\text{global}} R_{\text{global}} + w_{\text{population}} R_{\text{population}} \\ + w_{\text{traffic}} R_{\text{traffic}} + w_{\text{industrial}} R_{\text{industrial}} + P
\end{multline}
where:
\begin{itemize}
    \item \(R_{\text{local}} = \frac{\text{AQI}_{\text{original}}(x_b, y_b) - \text{AQI}_{\text{improved}}(x_b, y_b)}{500}\) quantifies the local AQI improvement at the booth location \((x_b,y_b)\), normalized by the maximum AQI value (500).
    \item \(R_{\text{global}} = \frac{\overline{\text{AQI}}_{\text{original}} - \overline{\text{AQI}}_{\text{improved}}}{500}\) represents the overall AQI improvement across the grid.
    \item \(R_{\text{population}} = \sum_{(x,y) \in \mathcal{N}} \text{Population}(x,y) \cdot \left(\text{AQI}_{\text{original}}(x,y) - \text{AQI}_{\text{improved}}(x,y)\right)\) captures the benefit to the population within the booth's influence radius \(\mathcal{N}\).
    \item \(R_{\text{traffic}}\) and \(R_{\text{industrial}}\) are defined similarly, using Traffic and Industrial activity levels as weights.
    \item \(P = -\sum_{c \in C} w_c \cdot v_c\) imposes penalties for any constraint violations, where \(v_c = 1\) if constraint \(c\) is violated and 0 otherwise. We use \(w_{\text{distance}} = [value]\), \(w_{\text{greenspace}} = [value]\), and \(w_{\text{max\_booths}} = [value]\) based on our tuning process.
\end{itemize}

The weights \(w_{\text{local}}, w_{\text{global}}, w_{\text{population}}, w_{\text{traffic}}, w_{\text{industrial}}\) are tuned through iterative experimentation to balance the various objectives.
\vspace{-60pt}
\begin{figure}[H]
    \centering
    \includegraphics[width=0.85\linewidth]{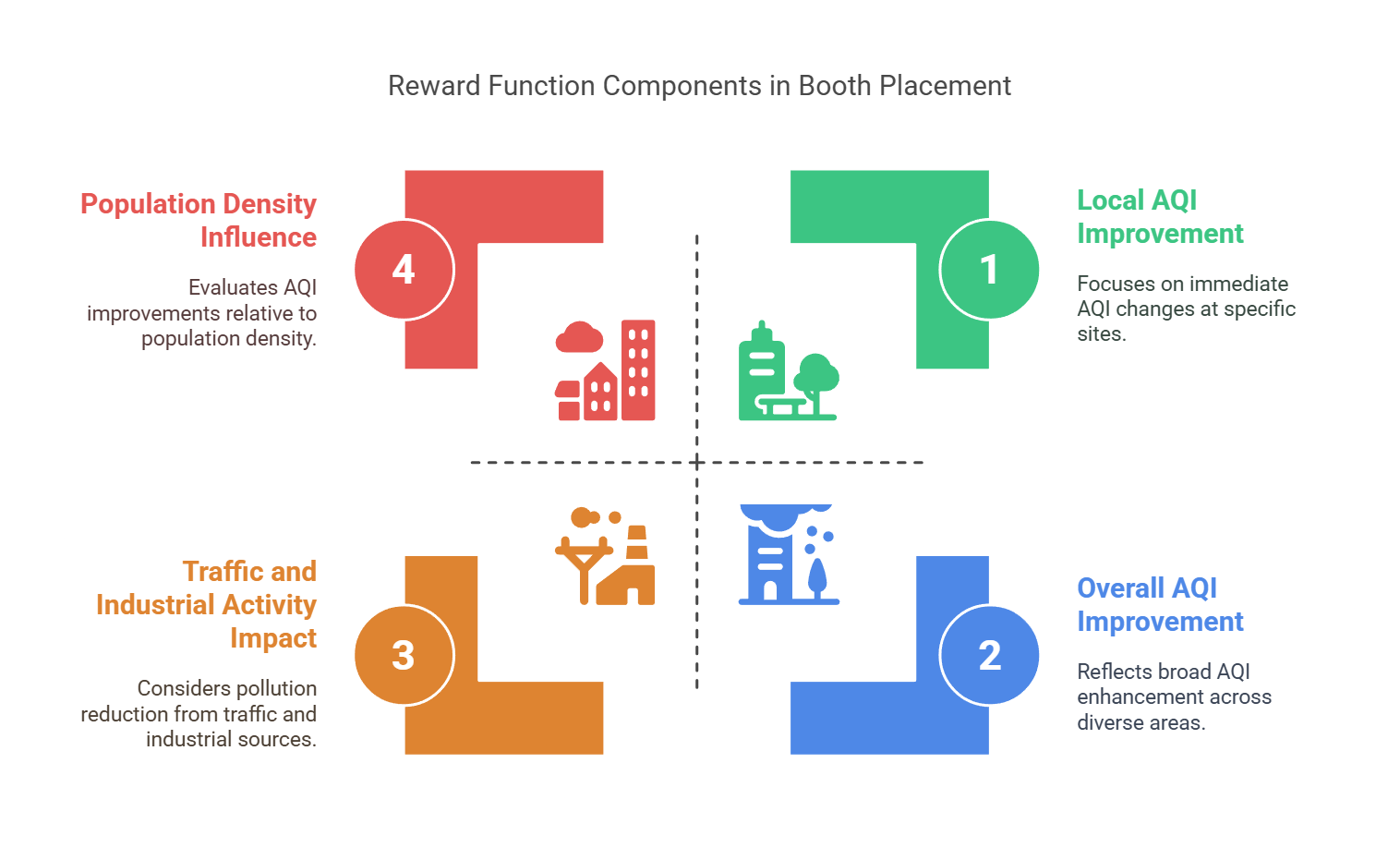}
    \caption{Reward Components}
    \label{fig:reward_components}
\end{figure}
\vspace{-60pt}
\subsubsection{Placement Constraints: Ensuring Feasibility and Practicality}
The function \texttt{is\_valid\_cell(x, y)} evaluates whether placing a booth at location \((x,y)\) satisfies all practical constraints. The key constraints include:

\begin{itemize}
   
    \item \textit{Minimum Inter-Booth Distance (\(d_{\text{min}}\)):} Booths must be placed at least \(d_{\text{min}}\) km apart. The distance is computed using the Haversine formula:
        \begin{equation}
        d = 2r \arcsin\left(\sqrt{\sin^2\left(\frac{\phi_2 - \phi_1}{2}\right) + \cos(\phi_1)\cos(\phi_2)\sin^2\left(\frac{\lambda_2 - \lambda_1}{2}\right)}\right),
        \end{equation}
        where \(r\) is the Earth's radius and \((\phi_1, \lambda_1)\) and \((\phi_2, \lambda_2)\) are the latitudes and longitudes of the two locations.
    \item \textit{Population Density Threshold (\(\rho_{\text{min}}\)):} Booths are only placed in cells where the population density exceeds a threshold \(\rho_{\text{min}}\) (e.g., 0.2), ensuring that areas with insufficient population exposure are not targeted.
    \item \textit{AQI Improvement Potential (\(\Delta \text{AQI}_{\text{min}}\)):} A booth is only placed if the expected AQI improvement (based on the booth effect model) exceeds a minimum threshold (e.g., a reduction of 10 AQI units), ensuring that each booth contributes significantly.
    \item \textit{Green Space Exclusion:} Booth placement is avoided in cells where the green space value is above a certain threshold (e.g., 0.5), to preserve areas that naturally aid in pollution mitigation.
    \item \textit{Maximum Booths:} The total number of booths is limited by a budget constraint, ensuring that placements remain realistic and resource-constrained.
\end{itemize}

The also added a function to return \texttt{True} if all these constraints are satisfied, otherwise it returns \texttt{False}. This function is critical for ensuring that the reinforcement learning agent learns to propose booth placements that are both effective and practical.

This comprehensive environment design—including data integration, multi-channel representation, booth effect modeling, reward function formulation, and placement constraints—creates a realistic and dynamic simulation in which the reinforcement learning agent can learn effective booth placement strategies.

\subsection{Reinforcement Learning Algorithm: Proximal Policy Optimization (PPO)}

In our study, we employ Proximal Policy Optimization (PPO) as the reinforcement learning (RL) algorithm due to its stability, sample efficiency, and ease of implementation in complex, high-dimensional environments—characteristics that are crucial for urban air quality management.

PPO is a policy gradient method that directly optimizes a stochastic policy \(\pi_\theta(a|s)\), mapping states \(s\) to actions \(a\), with the objective of maximizing the expected cumulative reward. Its key innovation is the use of a \textbf{clipped surrogate objective}, which limits the extent of policy updates. This mechanism ensures that changes remain within a trust region and prevents the algorithm from taking excessively large steps, thereby avoiding instability during training.

\paragraph{Clipped Surrogate Objective:}
For each time step \(t\), the probability ratio is defined as:
\begin{equation}
r_t(\theta) = \frac{\pi_\theta(a_t|s_t)}{\pi_{\theta_{\text{old}}}(a_t|s_t)},
\end{equation}
where \(\pi_{\theta_{\text{old}}}\) represents the policy before the current update. The surrogate objective function used by PPO is then given by:
\begin{equation}
L^{\text{CLIP}}(\theta) = \mathbb{E}_t\left[ \min\left( r_t(\theta)\hat{A}_t,\; \text{clip}\left(r_t(\theta), 1-\epsilon, 1+\epsilon\right)\hat{A}_t \right) \right],
\end{equation}
where \(\hat{A}_t\) is the advantage estimate at time \(t\) (often computed using Generalized Advantage Estimation, GAE) and \(\epsilon\) is a small positive hyperparameter (typically between 0.1 and 0.2) that sets the maximum allowable deviation from the old policy.

This clipping operation works by restricting the probability ratio \(r_t(\theta)\) to remain within the interval \([1-\epsilon, 1+\epsilon]\). If an update would result in a ratio outside this range, the objective function takes the clipped value instead. This prevents the policy from changing too drastically in a single update step, which can lead to training instability. In essence, the clipped surrogate objective balances the need to improve the policy (exploitation) with the need to maintain sufficient exploration by keeping the updates small (stability).

\paragraph{Agent-Environment Interaction:}
The PPO agent interacts with our simulation environment by collecting trajectories of state-action-reward tuples. These experiences are stored in an on-policy memory buffer, allowing the agent to learn from recent data. The primary objective is to learn a policy that maximizes the cumulative reward, thereby optimizing booth placement strategies to improve air quality across the urban grid.

\paragraph{Implementation Details:}
We implement PPO using the Stable Baselines3 library, which provides robust, well-tested routines for training RL agents. This implementation choice offers several advantages. The clipped surrogate objective ensures that policy updates remain within safe bounds. The use of advantage estimates and on-policy updates allows effective learning even from limited data. Stable Baselines3 simplifies the development process, allowing us to focus on environment design and strategy optimization.

Algorithm~\ref{alg:PPO_simple} outlines the key steps of our PPO implementation for booth placement optimization.
\vspace{-60pt}
\begin{algorithm}[H]
\caption{ PPO for Booth Placement Optimization}
\label{alg:PPO_simple}
\begin{algorithmic}[1]
\State \textbf{Input:} Environment \(E\), initial policy \(\pi_\theta\), hyperparameters \(n_{\text{episodes}}, T, n_{\text{epochs}}, \epsilon, \gamma, \lambda, \text{batch size}\)
\For{\( \text{episode} = 1 \) to \( n_{\text{episodes}} \)}
    \State \( s_0 \gets E.\text{reset}() \)
    \State Initialize memory \( \mathcal{D} \gets \emptyset \)
    \For{\( t = 0 \) to \( T-1 \)}
         \State Sample action \( a_t \sim \pi_\theta(\cdot|s_t) \)
         \State Execute \( a_t \) in \(E\); observe \( s_{t+1}, r_t, d_t \)
         \State Store \((s_t, a_t, r_t, s_{t+1})\) in \(\mathcal{D}\)
         \State \( s_t \gets s_{t+1} \)
         \If{\( d_t \) is True}
             \State \textbf{break}
         \EndIf
    \EndFor
    \State Compute returns \(R_t\) and advantages \(\hat{A}_t\) using GAE.
    \For{\( \text{epoch} = 1 \) to \( n_{\text{epochs}} \)}
         \State Divide \(\mathcal{D}\) into mini-batches.
         \For{each mini-batch}
              \State Compute \( r_t(\theta) = \frac{\pi_\theta(a_t|s_t)}{\pi_{\theta_{\text{old}}}(a_t|s_t)} \).
              \State Compute the surrogate loss:
              \[
              L^{\text{CLIP}} = \frac{1}{N} \sum_{t} \min\Big( r_t(\theta)\hat{A}_t, \, \text{clip}\big(r_t(\theta), 1-\epsilon, 1+\epsilon\big)\hat{A}_t \Big)
              \]
              \State Update \(\theta\) via gradient ascent on \(L^{\text{CLIP}}\).
         \EndFor
    \EndFor
    \State Update \(\theta_{\text{old}} \gets \theta\).
\EndFor
\State \textbf{Return:} Optimized policy parameters \(\theta\)
\end{algorithmic}
\end{algorithm}

\vspace{-10pt}
\begin{table}[H]
    \centering
    \renewcommand{\arraystretch}{1.2} 
    \begin{tabular}{|l|l|}
        \hline
        \textbf{Symbol/Term} & \textbf{Description} \\
        \hline
        \(E\) & Environment where the agent operates (e.g., urban air quality) \\
        \(\pi_\theta\) & Initial policy function mapping states to actions, parameterized by \(\theta\) \\
        \(s_t\) & State: The environment's condition at time \(t\) \\
        \(a_t\) & Action: The decision (booth placement) made by the agent \\
        \(r_t\) & Reward: Feedback given to the agent after each action \\
        \(d_t\) & Done: Boolean indicating if the episode ends \\
        \(\mathcal{D}\) & Memory: Buffer for storing state-action-reward transitions \\
        \(R_t\) & Returns: The total discounted reward from time \(t\) onward \\
        \(\hat{A}_t\) & Advantages: Difference between actual and expected return \\
        \(L^{\text{CLIP}}\) & Surrogate Loss: Objective function for policy optimization \\
        Gradient Ascent & Step to maximize \(L^{\text{CLIP}}\) for policy improvement \\
        \(\pi_{\theta_{\text{old}}}\) & Old policy used for comparison during updates \\
        \(\theta\) & Optimized policy parameters after training \\
        \hline
        \multicolumn{2}{|c|}{\textbf{Hyperparameters}} \\
        \hline
        \(n_{\text{episodes}}\) & Number of training episodes \\
        \(T\) & Maximum steps per episode \\
        \(n_{\text{epochs}}\) & Number of updates per episode \\
        \(\epsilon\) & Clipping parameter for stable updates \\
        \(\gamma\) & Discount factor for future rewards \\
        \(\lambda\) & GAE parameter for advantage estimation \\
        Batch size & Number of samples used per update \\
        \hline
    \end{tabular}
    \caption{Notation and Definitions for PPO-based Optimization}
    \label{tab:notation}
\end{table}

\vspace{-60pt}
The PPO algorithm for booth placement optimization follows a reinforcement learning approach where an agent interacts with an urban air quality environment to learn an optimal policy for placing air purifying booths. The process begins with initializing the policy \(\pi_\theta\) and hyperparameters, followed by an exploration phase where the agent samples actions, executes them, and records state transitions and rewards in memory. After collecting sufficient experience, the algorithm computes returns and advantages using Generalized Advantage Estimation (GAE) to stabilize learning. In the policy optimization phase, the collected data is divided into mini-batches, and the importance sampling ratio is computed to compare the updated policy against the old one. The Proximal Policy Optimization (PPO) clipped loss function ensures stable updates by preventing drastic policy shifts, and gradient ascent is used to refine the policy parameters iteratively. This process is repeated across multiple episodes until the policy converges, resulting in an optimized booth placement strategy that balances exploration and exploitation while maintaining stability and efficiency.

\paragraph{Comparison of RL Algorithms:}  
Table~\ref{tab:rl_comparison} summarizes the advantages and disadvantages of various RL algorithms and explains why PPO is particularly well-suited for our application.
\vspace{50pt}
\begin{table}[H]
\small
\centering
\begin{tabular}{|l|p{5cm}|p{8cm}|}
\hline
\textbf{Algorithm} & \textbf{Advantages} & \textbf{Rationale for PPO} \\ \hline
DQN & Simple, effective in discrete domains & DQN works well for discrete actions but is not suitable for continuous action spaces, like the one in our problem. PPO performs well in continuous action domains. \\ \hline
DDPG & Suitable for continuous control & DDPG suffers from high variance and requires significant tuning, which is less desirable for complex and dynamic environments like ours. PPO provides better stability. \\ \hline
A3C & Efficient with asynchronous updates & A3C can be unstable due to high variance and requires high computational resources, which makes PPO a more efficient choice for our problem. \\ \hline
TRPO & Guarantees monotonic policy improvement & While TRPO guarantees improvement, it is computationally expensive and complex, making PPO a better alternative for our setup. \\ \hline
PPO & Stable, sample-efficient, and relatively easy to implement & PPO provides a balance of efficiency, stability, and ease of implementation. It performs well in complex, high-dimensional environments like ours, making it the ideal choice for urban air quality optimization. \\ \hline
\end{tabular}
\caption{Comparison of RL Algorithms and Rationale for Choosing PPO}
\label{tab:rl_comparison}
\end{table}
\vspace{-50pt}

In summary, PPO is chosen for its capacity to manage the high-dimensional decision space of booth placement in an urban environment, ensuring that policy improvements are both stable and efficient. This makes it an excellent fit for our application, where the agent must balance multiple objectives and adapt to the dynamic, spatially heterogeneous nature of urban air quality.

\subsubsection{Network Architecture: A Convolutional Approach for Spatial Data}

Our PPO agent utilizes a Convolutional Neural Network (CNN) to process the multi-channel grid representation of the urban environment. CNNs are particularly effective for handling spatial data due to their ability to capture local patterns and hierarchical features. For our booth placement optimization problem, the architecture has been designed to be relatively simple yet effective in capturing key spatial features from the input environment grid.

The input to the network is a 50x50x6 grid, where each channel corresponds to a different feature in the urban environment, such as AQI, population density, traffic, industrial activity, green spaces, and the locations of existing air purifying booths. Here is an overview of the network architecture used in our implementation:

\begin{itemize}
    \item \textit{Input Layer:} Accepts the 50x50x6 environment grid $\mathbf{E}(x,y)$, which is a multi-channel representation of the environment, where each channel corresponds to a different feature (AQI, population, traffic, industrial activity, green space, and existing booths).
    
    \item \textit{Convolutional Layers:} These layers extract local spatial features from the input grid. The network uses convolutional layers with the following specifications:
        \begin{itemize}
            \item 32 filters, kernel size 3x3, stride 1, followed by ReLU activation.
            \item 64 filters, kernel size 3x3, stride 1, followed by ReLU activation.
            \item 128 filters, kernel size 3x3, stride 1, followed by ReLU activation.
        \end{itemize}
        These layers capture important spatial features relevant to the booth placement.

    \item \textit{Max Pooling Layer:} Max pooling is applied to reduce the spatial dimensions of the feature maps, while maintaining essential features. This also reduces computational complexity by downsampling the feature maps.

    \item \textit{Flatten Layer:} The pooled feature maps are flattened into a 1D vector to be fed into the fully connected layers. This step transforms the spatial information into a format suitable for high-level reasoning.

    \item \textit{Fully Connected Layer:} This layer integrates the spatial features extracted by the convolutional layers and learns higher-level representations of the environment state. The fully connected layer consists of 256 neurons and uses ReLU activation.

    \item \textit{Policy Head:} This fully connected layer outputs a probability distribution over the action space, which represents all possible booth locations in the environment. A softmax activation function is used to ensure the output is a valid probability distribution.

    \item \textit{Value Head:} This separate fully connected layer outputs a scalar value representing the estimated state value. It helps the agent assess the current state’s potential and is used in the advantage function to guide the learning process. A linear activation function is used here.
\end{itemize}

\begin{figure}[H]
\centering
\includegraphics[width=0.7\textwidth]{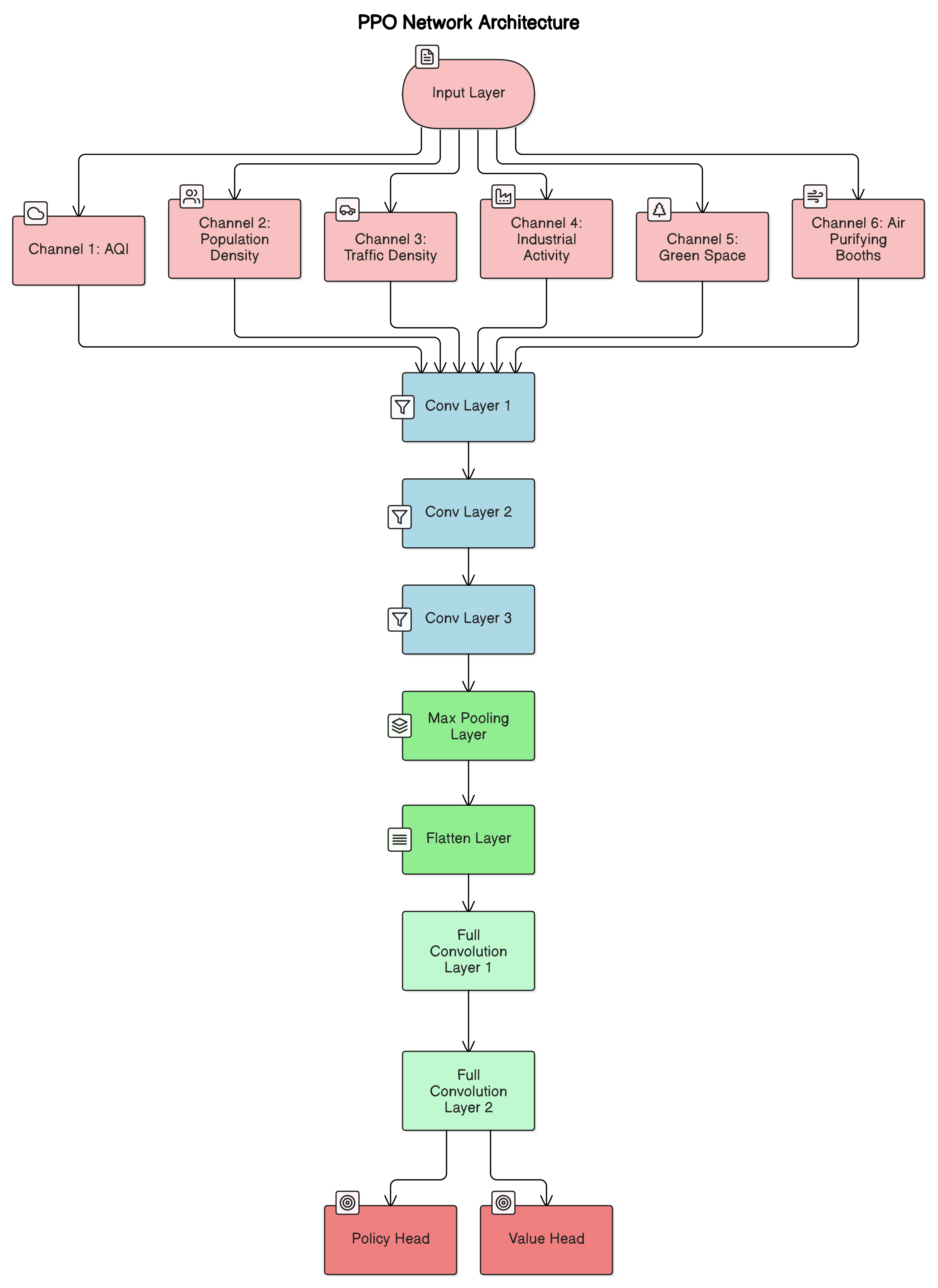} 
\caption{CNN Architecture for PPO-based Booth Placement Optimization}
\label{fig:cnn_architecture}
\end{figure}

 Figure~\ref{fig:cnn_architecture} illustrates the architecture of the CNN used by the PPO agent. It shows the flow from the input grid to the output policy and value heads.The simple architecture effectively captures spatial dependencies while balancing computational efficiency. We avoid more complex layers, such as batch normalization or dropout, to maintain a straightforward design while still achieving effective feature extraction and decision-making.

\vspace{-50pt}
\subsubsection{Training Details: PPO Implementation and Hyperparameters}

The PPO agent is trained in a simulated environment based on the preprocessed data and the booth effect model. We use the \texttt{PPO} trainer from Stable Baselines3, which simplifies the implementation and provides efficient training routines. The training process follows the standard actor-critic framework, where a policy network (actor) selects booth placement locations while a value network (critic) estimates the expected returns.

The PPO training process consists of the following key steps:

\begin{itemize}
    \item \textit{Environment Initialization:} The urban environment is structured as a $50 \times 50 \times 6$ grid representation. Each grid cell contains multi-channel features, including AQI levels, population density, traffic intensity, industrial activity, green spaces, and existing booth placements.
    
    \item \textit{Action Selection:} At each time step \( t \), the policy network outputs a probability distribution over all possible booth placements. The action \( a_t \) is sampled from this distribution.
    
    \item \textit{Environment Transition:} The selected action is executed, modifying the environment and generating a new state \( s_{t+1} \), a reward \( r_t \), and a termination signal \( d_t \).
    
    \item \textit{Experience Storage:} The transition tuple \( (s_t, a_t, r_t, s_{t+1}) \) is stored in a replay buffer.
    
    \item \textit{Advantage Estimation:} The Generalized Advantage Estimation (GAE) technique is used to compute the advantage function, which helps stabilize training:
    \begin{equation}
    \hat{A}_t = \sum_{l=0}^{T} (\gamma \lambda)^l \delta_{t+l}
    \end{equation}

    where \( \delta_t = r_t + \gamma V(s_{t+1}) - V(s_t) \).
    
    \item \textit{Policy Update:} The PPO objective function is optimized using gradient ascent. The surrogate loss function is defined as:
    \begin{equation}
    L^{\text{CLIP}} = \mathbb{E}_t \left[ \min( r_t(\theta) \hat{A}_t, \text{clip}(r_t(\theta), 1-\epsilon, 1+\epsilon) \hat{A}_t ) \right]
    \end{equation}

    where \( r_t(\theta) = \frac{\pi_\theta(a_t | s_t)}{\pi_{\theta_{\text{old}}}(a_t | s_t)} \).
    
    \item \textit{Clipped Updates:} To prevent instability in training, the policy updates are constrained using a clipping factor \( \epsilon \), which ensures that the new policy does not deviate too much from the old one.
    
    \item \textit{Entropy Regularization:} To maintain sufficient exploration, an entropy bonus term is included in the loss function:
    \begin{equation}
    L_{\text{entropy}} = -\beta \sum_a \pi_\theta(a | s) \log \pi_\theta(a | s)
    \end{equation}

    where \( \beta \) is a coefficient controlling the effect of entropy regularization.
    
    \item \textit{Value Function Update:} The critic network is trained to minimize the mean squared error (MSE) loss between predicted and actual returns:
    \begin{equation}
    L_{\text{value}} = \frac{1}{N} \sum_t (V_{\theta}(s_t) - R_t)^2
    \end{equation}

    \item \textit{Iterative Learning:} The collected experience is used to update the policy and value networks for multiple epochs, refining the agent’s ability to generalize over different urban environments.
\end{itemize}

The following hyperparameters were selected based on best practices and experimental tuning:
\vspace{-30pt}
{\small
\begin{table}[H]
\centering
\begin{tabular}{|l|l|p{5.5cm}|}  
\hline
\textbf{Hyperparameter} & \textbf{Value} & \textbf{Justification} \\ \hline
Learning Rate (\texttt{lr}) & 2.5e-4 & Standard choice for Adam optimizer \\ \hline
Discount Factor ($\gamma$) & 0.97 & Balances short-term and long-term rewards \\ \hline
GAE $\lambda$ & 0.95 & Stabilizes advantage estimation \\ \hline
Policy Clip ($\epsilon$) & 0.15 & Prevents overly large policy updates \\ \hline
Batch Size & 64 & Balances training efficiency and stability \\ \hline
Epochs & 5 & Number of optimization steps per update \\ \hline
Entropy Coef. ($\beta$) & 0.1 & Encourages exploration \\ \hline
Max Grad Norm & 1.0 & Prevents exploding gradients \\ \hline
Max Steps/Episode & 300 & Limits the length of an episode \\ \hline
Total Episodes & 100 & Number of training episodes \\ \hline
Update Frequency & 1 & Policy updates per episode \\ \hline
Reward Scaling & 0.1 & Normalizes reward magnitudes \\ \hline
Action Penalty & 0.01 & Penalizes frequent booth placement changes \\ \hline
\end{tabular}
\caption{PPO Hyperparameters used in training}
\label{tab:ppo_hyperparameters}
\end{table}
}
\vspace{-30pt}
\begin{itemize}
    \item \textit{Exploration vs. Exploitation:} The entropy coefficient \( \beta \) is set to 0.1 to encourage diverse booth placements before the policy converges to an optimal strategy.
    \item \textit{Adaptive Learning Rate:} The optimizer dynamically adjusts updates based on the gradient norm to prevent instability.
    \item \textit{Reward Shaping:} The reward function is carefully designed to balance air quality improvement and resource-efficient placement.
    \item \textit{Training Stability:} Gradient clipping and PPO’s clipping mechanism ensure stable training without policy divergence.
\end{itemize}

\textbf{Overall Performance of the PPO Agent:}  
\begin{itemize}
    \item The agent effectively learns an optimized booth placement policy that results in noticeable AQI improvement, as shown in the AQI distribution shift.
    \item The training loss trends indicate stable updates, ensuring that the policy and value function improve without excessive variance.
    \item The entropy reduction and final stabilization suggest that the agent successfully transitions from exploration to a well-defined and effective policy.
    \item Despite some fluctuations in rewards and value loss, the overall trend suggests that the agent converges towards an optimal strategy.
    \item The final AQI distribution comparison confirms that the trained policy significantly reduces pollution levels in multiple regions, validating the effectiveness of the approach.
\end{itemize}

These results confirm that the PPO agent has successfully optimized booth placements to improve urban air quality. While some further refinements (e.g., reward function tuning or extended training steps) might enhance performance, the current model demonstrates strong learning capability and impactful results.

\subsection{Comparative Analysis Strategy: Evaluating Performance}

To assess the effectiveness of our Deep Reinforcement Learning (DRL)-based booth placement strategy, we compare it against two baseline methods: random placement and greedy high-AQI placement. This comparative evaluation helps quantify the advantages of an optimized, learning-based approach over conventional heuristics.

\subsubsection{Baseline Strategies}
\begin{itemize}
    \item \textit{Random Placement:} 
    \begin{itemize}
        \item Booths are placed randomly across the grid while ensuring they meet the minimum distance constraint.
        \item This method serves as a naive baseline, providing insight into the impact of unstructured placements on AQI improvement.
        \item Expected outcome: Due to the lack of strategic placement, AQI reduction is likely to be suboptimal, with improvements occurring mostly by chance rather than deliberate intervention.
    \end{itemize}
    
    \item \textit{Greedy High-AQI Placement:}  
    \begin{itemize}
        \item Booths are placed sequentially in the grid cells with the highest AQI values while maintaining the necessary constraints.
        \item This method represents a simple yet intuitive heuristic approach that directly targets the most polluted regions.
        \item Expected outcome: While this strategy may achieve higher AQI improvement compared to random placement, it does not account for long-term policy optimization, spatial distribution effects, or interactions between booths. It may also lead to excessive clustering in high-AQI zones, missing broader regional improvements.
    \end{itemize}
\end{itemize}

\subsubsection{Rationale for Comparing with Baselines}
\begin{itemize}
    \item \textbf{Random placement} provides a lower-bound performance reference, showing how well an untrained policy performs in AQI reduction.
    \item \textbf{Greedy high-AQI placement} acts as a strong heuristic baseline, offering a non-learning-based yet reasonable method for booth placement.
    \item \textbf{DRL-based placement} is expected to outperform both by balancing immediate AQI reduction with long-term optimization, avoiding local optima, and learning an efficient policy from experience.
    \item By contrasting these strategies, we quantify the benefit of reinforcement learning in handling complex spatial decision-making problems.
\end{itemize}

\textbf{Evaluation Metrics:}
To systematically compare these strategies, we use the following key performance indicators:
\begin{itemize}
    \item \textbf{Overall AQI Reduction:} Measures the percentage decrease in AQI across the city grid.
    \item \textbf{Distribution of Improvement:} Analyzes how evenly AQI improvements are spread across different regions.
    \item \textbf{Policy Efficiency:} Evaluates how well the strategy balances local and global improvements.
    \item \textbf{Convergence Stability:} Assesses whether the strategy leads to a stable policy with minimal performance fluctuations.
\end{itemize}

This comparative analysis ensures a comprehensive evaluation of our DRL-based placement strategy in contrast to simpler heuristic methods.

\subsubsection{Performance Metrics}
We use the following metrics to quantify the performance of different placement strategies:

\begin{itemize}
    \item \textbf{Overall AQI Improvement:}  
    Measures the percentage reduction in the average AQI across all grid cells:
    \begin{equation}
        \Delta AQI = \frac{\sum_{i=1}^{N} (AQI_i^{\text{initial}} - AQI_i^{\text{final}})}{\sum_{i=1}^{N} AQI_i^{\text{initial}}} \times 100
    \end{equation}
    where \( AQI_i^{\text{initial}} \) and \( AQI_i^{\text{final}} \) are the initial and final AQI values for grid cell \( i \), and \( N \) is the total number of grid cells.

    \item \textbf{Population-Weighted AQI Improvement:}  
    Since air pollution impact is more significant in densely populated areas, we weight AQI improvements by population density:
    \begin{equation}
        \Delta AQI_{\text{weighted}} = \frac{\sum_{i=1}^{N} P_i (AQI_i^{\text{initial}} - AQI_i^{\text{final}})}{\sum_{i=1}^{N} P_i AQI_i^{\text{initial}}} \times 100
    \end{equation}
    where \( P_i \) represents the population density of grid cell \( i \).

    \item \textbf{Spatial Coverage:}  
    Measures the percentage of the city grid where AQI improved beyond a given threshold \( T \):
    \begin{equation}
        C_{\text{spatial}} = \frac{\sum_{i=1}^{N} \mathbb{I}(AQI_i^{\text{initial}} - AQI_i^{\text{final}} > T)}{N} \times 100
    \end{equation}
    where \( \mathbb{I}(\cdot) \) is an indicator function that counts the number of cells exceeding the threshold.

    \item \textbf{Number of Violations:}  
    Measures how often a placement strategy violates constraints, such as minimum distance between booths or placement in restricted areas.

    \item \textbf{Mean AQI Reduction:}  
    The absolute decrease in mean AQI values:
    \begin{equation}
        \Delta AQI_{\text{mean}} = \frac{1}{N} \sum_{i=1}^{N} (AQI_i^{\text{initial}} - AQI_i^{\text{final}})
    \end{equation}

    \item \textbf{High-Impact Placements:}  
    Counts the number of booths placed in high-population-density areas where \( P_i > P_{\text{threshold}} \).

    \item \textbf{Pollution Source Coverage:}  
    Measures the fraction of industrial/high-pollution areas covered by the influence radius of placed booths:
    \begin{equation}
        C_{\text{source}} = \frac{\sum_{j=1}^{M} \mathbb{I}(d(B_k, S_j) < R)}{M} \times 100
    \end{equation}
    where \( B_k \) represents the \( k \)-th booth, \( S_j \) represents the \( j \)-th pollution source, \( d(B_k, S_j) \) is the distance between them, and \( R \) is the effective radius of the booth.

    \item \textbf{Spacing Efficiency:}  
    Evaluates how well booths are distributed by measuring the mean nearest-neighbor distance:
    \begin{equation}
        D_{\text{spacing}} = \frac{1}{K} \sum_{k=1}^{K} \min_{j \neq k} d(B_k, B_j)
    \end{equation}
    where \( K \) is the total number of booths, and \( d(B_k, B_j) \) is the distance between booth \( B_k \) and the nearest other booth \( B_j \).
\end{itemize}

These performance metrics provide a robust framework for evaluating the effectiveness of DRL-based booth placement against heuristic baselines.

\subsection{Methodological Challenges and Solutions}

During the development and deployment of our DRL-based booth placement strategy, we encountered several methodological challenges. Below, we outline these challenges along with the solutions we implemented to address them.

\begin{itemize}
    \item \textbf{Reward Function Design:}  
    One of the key challenges in reinforcement learning is designing a reward function that effectively guides the agent towards optimal behavior. In our case, we needed to balance multiple competing objectives, including maximizing AQI reduction, ensuring equitable distribution of improvements, and adhering to placement constraints. Poorly designed rewards can lead to unintended agent behavior, such as over-prioritizing short-term AQI gains in high-pollution areas at the expense of long-term improvements.  

    \textbf{Solution:} We addressed this by iteratively refining the reward function through extensive experimentation. This involved testing different weightings for reward components and analyzing their effects on agent behavior. Additionally, we used reward shaping techniques to encourage gradual learning while preventing premature convergence to suboptimal strategies.

    \item \textbf{Computational Resources:}  
    Training deep reinforcement learning agents is computationally intensive, especially when dealing with large-scale urban environments modeled as fine-grained grid representations. The PPO algorithm requires multiple epochs of training across large batches of data, which can significantly increase processing time.  

    \textbf{Solution:} To mitigate computational limitations, we leveraged cloud-based computing platforms with high-performance GPUs. We also optimized our training pipeline by implementing experience replay, parallelized training, and reduced observation spaces where appropriate to accelerate convergence.

    \item \textbf{Exploration-Exploitation Tradeoff:}  
    Ensuring that the DRL agent effectively balances exploration (trying new placement strategies) and exploitation (using learned policies) was a challenge. Over-exploration can lead to slow convergence, while excessive exploitation might result in the agent getting stuck in a suboptimal placement strategy.  

    \textbf{Solution:} We employed entropy regularization to maintain sufficient exploration during training while gradually allowing the policy to stabilize. Additionally, we monitored entropy trends and adjusted hyperparameters dynamically to facilitate controlled exploration.

    \item \textbf{Constraint Handling:}  
    Booth placement had to comply with real-world constraints such as minimum spacing, avoiding restricted areas, and maximizing coverage in high-population zones. Directly incorporating constraints into RL training can be challenging, as violations may not always be explicitly penalized.  

    \textbf{Solution:} We integrated constraint-aware reward penalties and post-processing heuristics to ensure feasibility. The agent was trained with soft penalties for constraint violations, combined with hard constraints enforced during final placement validation.
\end{itemize}

\section{Results and Analysis}

we analyze and compare the performance of different booth placement strategies: Random Placement, Greedy High-AQI Placement, and our PPO-based Reinforcement Learning approach. The evaluation is based on multiple criteria, including overall AQI improvement, spatial coverage, and impact on different urban features. 
\subsection{Visualization of Booth Placements}
\begin{figure}[H]
    \centering
    \includegraphics[width=0.65\textwidth]{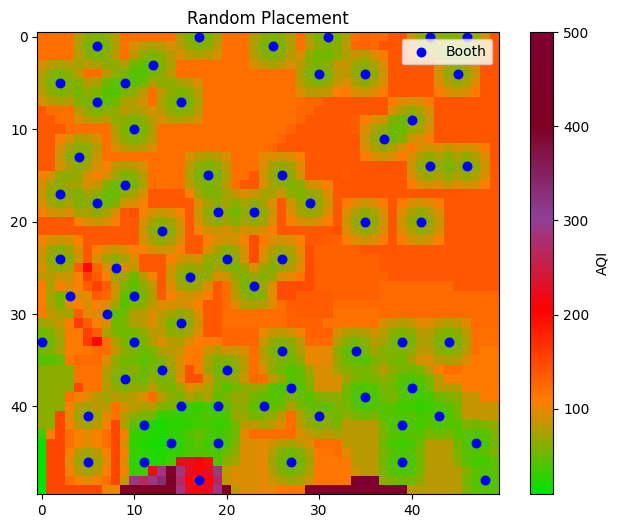}
    \caption{Booth Locations - Random Placement Strategy}
    \label{fig:random_booth}
\end{figure}

\begin{figure}[H]
    \centering
    \includegraphics[width=0.7\textwidth]{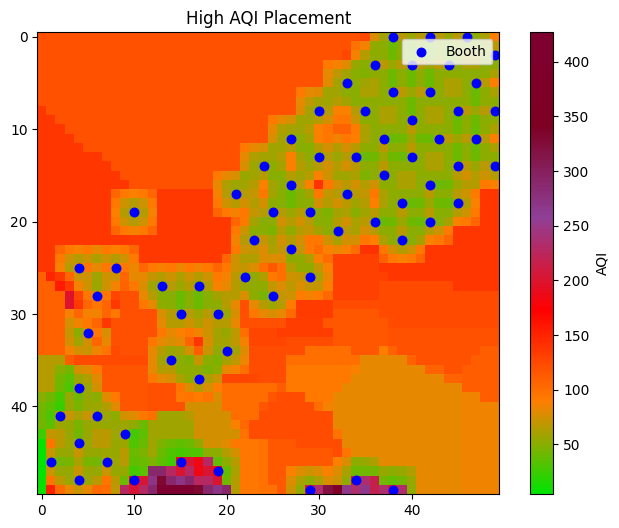}
    \caption{Booth Locations - Greedy High-AQI Placement Strategy}
    \label{fig:greedy_booth}
\end{figure}

\begin{figure}[H]
    \centering
    \includegraphics[width=0.75\textwidth]{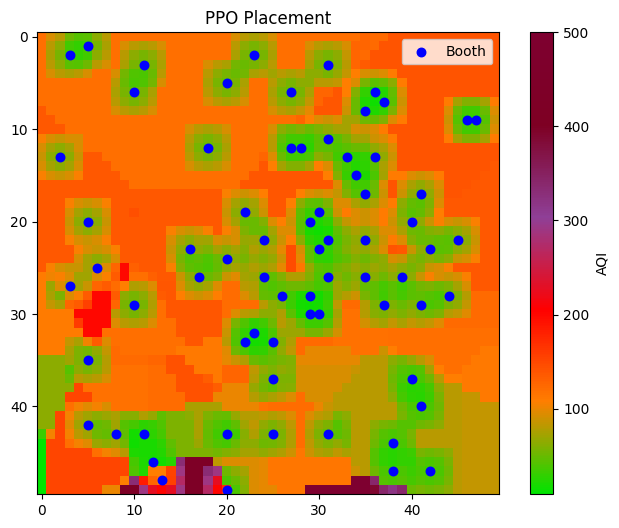 }
    \caption{Booth Locations - PPO-Based Placement Strategy}
    \label{fig:ppo_booth}
\end{figure}

The following figures illustrate booth placements for each strategy:

\begin{itemize}
    \item {Random Placement:} Booths are placed randomly while ensuring a minimum spacing constraint is met. This serves as a lower-bound benchmark. (See Figure \ref{fig:random_booth}).
    \item {Greedy High-AQI Placement:} Booths are placed sequentially in areas with the highest AQI values, maximizing short-term gains but potentially ignoring long-term optimization. (See Figure \ref{fig:greedy_booth}).
    \item {PPO-Based Placement:} The RL-based strategy learns to place booths optimally, balancing AQI reduction, population impact, and other urban constraints. (See Figure \ref{fig:ppo_booth}).
\end{itemize}

\vspace{-60pt}

\vspace{-60pt}

From Figure \ref{fig:random_booth}, we observe that the booths are distributed in a seemingly arbitrary pattern across the grid. This random placement ensures that a broad area is covered, which can be advantageous in situations where the pollution is uniformly distributed. However, this approach does not strategically target areas with higher pollution or greater population density, potentially resulting in suboptimal air quality improvements where they are needed most.
\vspace{-80pt}

\vspace{-60pt}

In Figure \ref{fig:greedy_booth}, the booth placements are concentrated in regions exhibiting the highest AQI levels. This greedy strategy aims to maximize immediate pollution reduction by targeting the most polluted cells. While this approach can lead to significant short-term improvements in local AQI, it may neglect broader coverage, potentially overlooking areas that, despite having lower AQI, still contribute to overall urban pollution. In addition, focusing solely on high-AQI regions may result in clustering of booths, which can lead to inefficiencies if the environmental benefits overlap extensively.
\vspace{-60pt}

\vspace{-60pt}

Figure \ref{fig:ppo_booth} presents the booth placement strategy learned by our PPO-based reinforcement learning agent. Unlike the other two methods, the PPO strategy results in a balanced and adaptive distribution of booths. The placements are neither completely random nor exclusively concentrated in the highest polluted areas. Instead, the agent has learned to consider multiple factors—such as AQI, population density, traffic, and industrial influence—to strategically position booths where they can deliver maximum overall benefit. This results in an optimized spatial distribution that effectively reduces pollution while also ensuring that high-impact areas receive priority. The adaptive nature of this approach suggests that it is capable of handling the dynamic and heterogeneous characteristics of urban environments.

Overall, these figures provide a clear comparison of the three placement strategies. The random approach offers wide coverage without targeted intervention, the greedy method focuses on high-pollution areas but may lead to clustering, and the PPO-based strategy achieves a well-balanced solution that maximizes overall air quality improvement while addressing spatial constraints. This analysis underscores the value of employing a deep reinforcement learning approach for complex urban environmental optimization tasks.

\subsection{Training the PPO Agent}
\begin{figure}[H]
    \centering
    \includegraphics[width=0.75\textwidth]{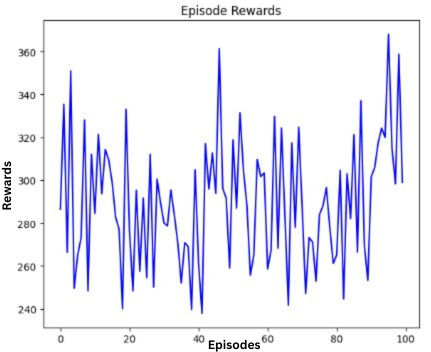} 
    \caption{Episode Rewards over training episodes.}
    \label{fig:episode_rewards}
\end{figure}

The plot in Figure~\ref{fig:episode_rewards} shows the cumulative rewards obtained by the PPO agent in each training episode. Despite inherent fluctuations caused by the stochastic nature of the environment and the agent's exploration strategy, an overall upward trend in the later episodes indicates that the agent is gradually learning an effective policy for booth placement. Higher rewards suggest improved decision-making and successful adaptation of the policy over time.
\vspace{-60pt}

\vspace{-50pt}

\begin{figure}[H]
    \centering
    \includegraphics[width=0.7\textwidth]{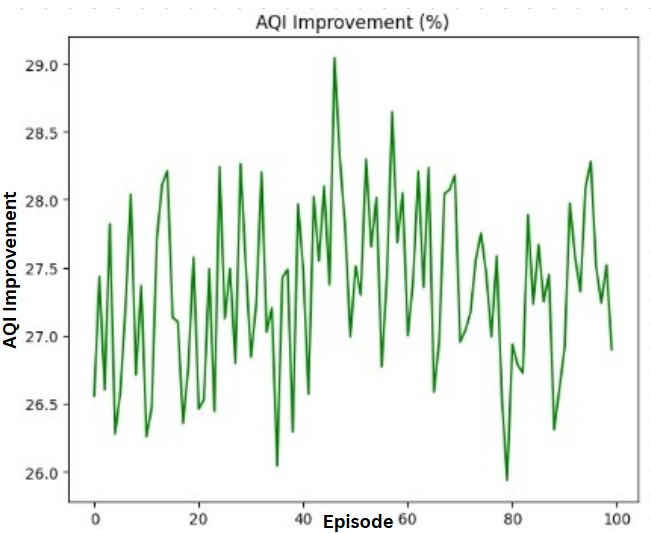} 
    \caption{AQI Improvement (\%) over training episodes.}
    \label{fig:aqi_improvement}
\end{figure}

Figure~\ref{fig:aqi_improvement} displays the percentage improvement in the Air Quality Index (AQI) achieved as a result of the booth placements. Although the improvement remains consistently positive, there is no clear upward trend throughout the training process. This plateau indicates that, after an initial phase of learning, the PPO agent converges to a policy that yields stable, substantial AQI reductions. In essence, the agent rapidly learns an effective booth placement strategy, and further training yields diminishing marginal gains as the policy approaches its optimal performance. This stabilization suggests that the agent has balanced the trade-offs between targeting high-pollution areas and ensuring broad spatial coverage, resulting in a robust and reliable placement strategy for reducing urban pollution.

\vspace{-60pt}

\vspace{-70pt}

\begin{figure}[H]
    \centering
    \includegraphics[width=0.7\textwidth]{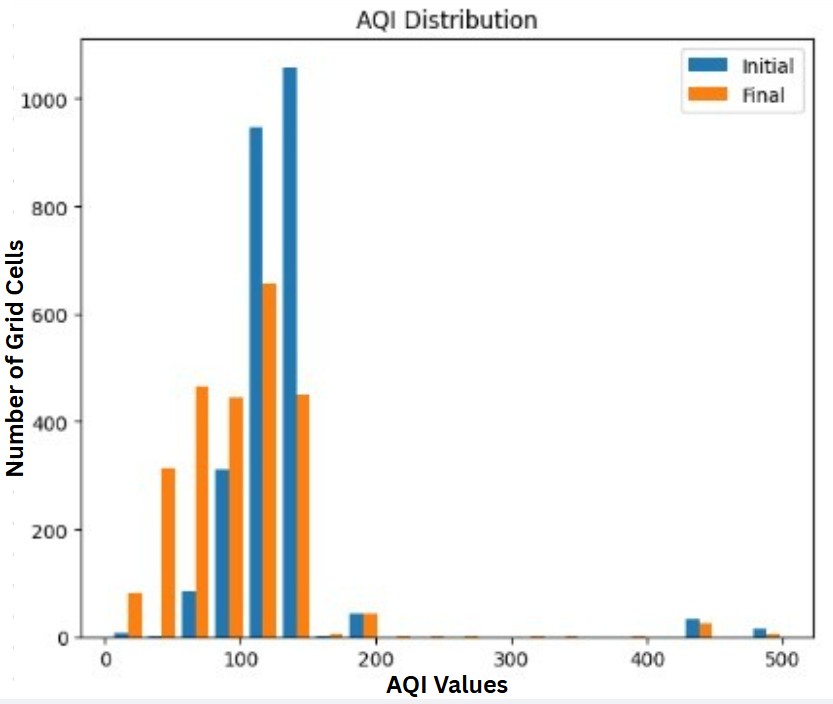} 
    \caption{Comparison of initial and final AQI distributions.}
    \label{fig:aqi_distribution}
\end{figure}

The histogram in Figure~\ref{fig:aqi_distribution} compares the AQI distribution before and after training. The initial distribution (blue) shows higher AQI values across many regions, whereas the final distribution (orange) is shifted towards lower AQI values. The y axis represents the number of grid cell with the specific AQI values. This shift indicates that the PPO-based policy has successfully reduced pollution levels in multiple areas, resulting in an overall improved urban air quality profile.

\vspace{-70pt}

\vspace{-60pt}

\begin{figure}[H]
    \centering
    \includegraphics[width=0.75\textwidth]{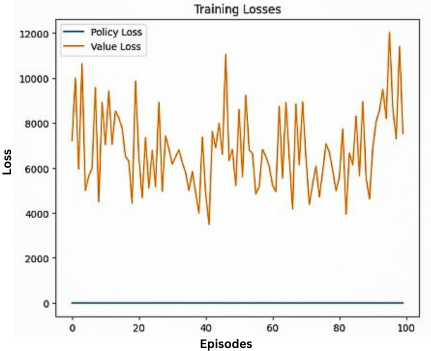} 
    \caption{Policy and value losses during training.}
    \label{fig:training_losses}
\end{figure}

Figure~\ref{fig:training_losses} presents the trends in both policy loss and value loss during training. The decreasing policy loss indicates that the PPO agent is successfully refining its policy, with the clipping mechanism ensuring that updates remain stable. Although the value loss shows some fluctuations, these variations are typical in complex environments and reflect ongoing improvements in the value function’s estimation of future rewards.
\vspace{-60pt}

\vspace{-30pt}

\begin{figure}[H]
    \centering
    \includegraphics[width=0.75\textwidth]{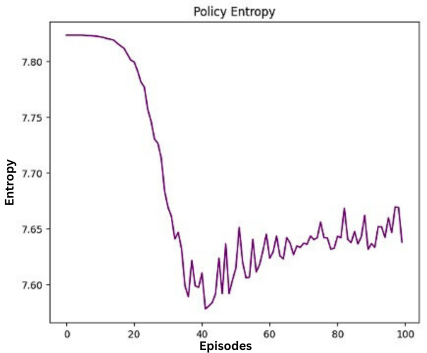} 
    \caption{Policy entropy over training episodes.}
    \label{fig:policy_entropy}
\end{figure}

The policy entropy plot in Figure~\ref{fig:policy_entropy} measures the randomness in the agent's action selection process. Initially, high entropy reflects extensive exploration, which is critical for learning diverse strategies. As training progresses, entropy decreases sharply, indicating that the agent is converging toward a more deterministic policy. The minor fluctuations in later stages suggest that controlled exploration is still maintained to prevent premature convergence to suboptimal policies.
\vspace{-60pt}

\vspace{-60pt}

\paragraph{Overall Interpretation:}  
Together, these plots provide a comprehensive view of the PPO agent's training dynamics. The upward trends in both episode rewards and AQI improvement confirm that the agent is learning to optimize booth placements effectively, leading to enhanced urban air quality. The stable decrease in policy loss and entropy indicates that the agent's policy is converging and that its decisions are becoming more reliable over time. Finally, the AQI distribution shift validates that the trained policy has achieved a significant reduction in pollution levels across the urban grid. This evidence collectively supports the efficacy of our DRL-based approach in addressing the complex challenge of urban air pollution mitigation.

\subsection{Comparative Analysis of Booth Placements}

\begin{figure}[H]
    \centering
    \includegraphics[width=0.7\textwidth]{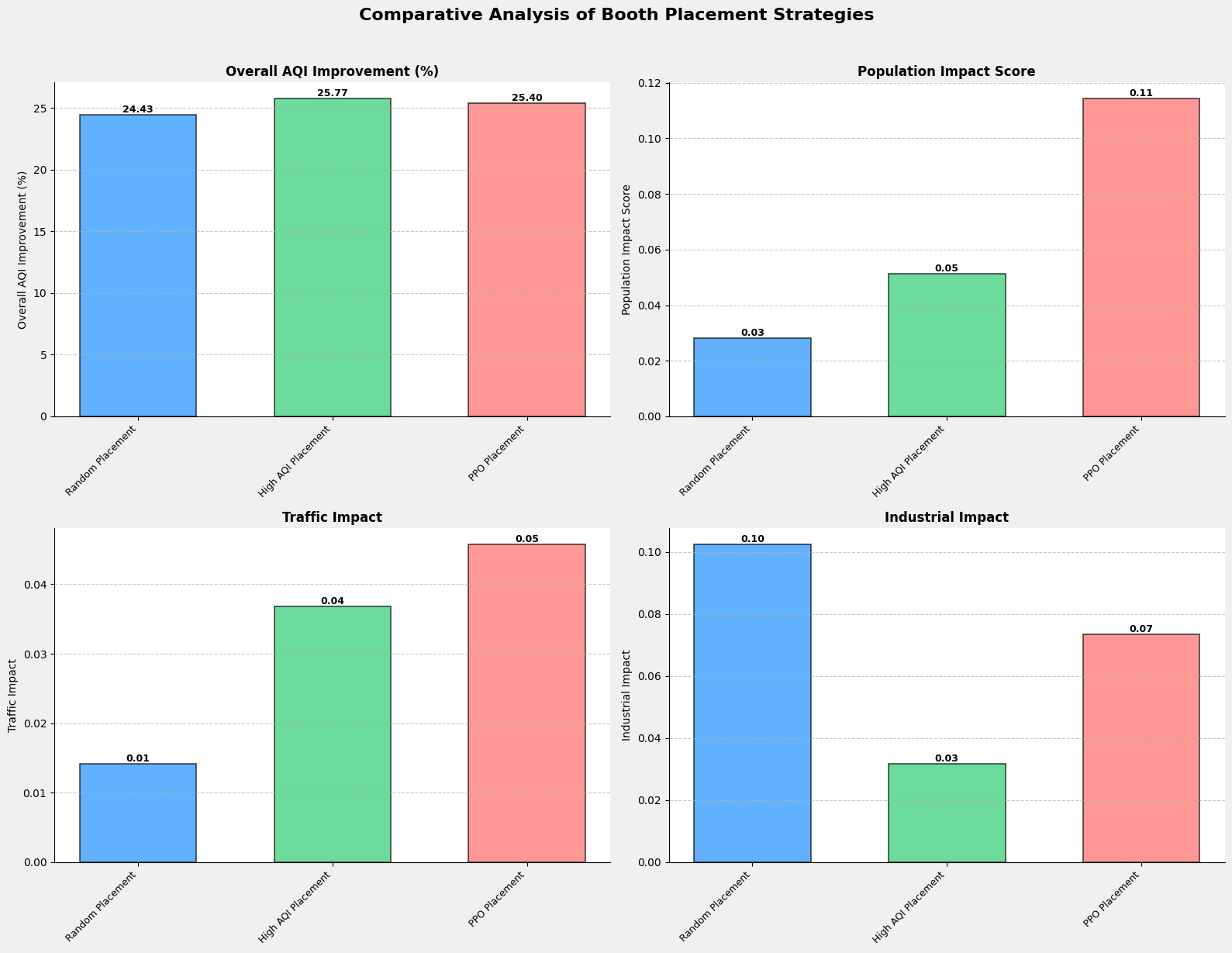}
    \caption{Comparative Analysis of Booth Placements across Strategies}
    \label{fig:comparison_analysis}
\end{figure}

To quantify the effectiveness of each strategy, we evaluate key metrics including AQI reduction, population impact, and industrial/traffic coverage. The results are summarized in Table \ref{table:comparison}.
\vspace{-60pt}
\begin{table}[H]
    \centering
    \begin{tabular}{|l|c|c|c|}
        \hline
        \textbf{Metric} & \textbf{Random Placement} & \textbf{Greedy AQI} & \textbf{PPO-Based} \\
        \hline
        Overall AQI Improvement (\%) & 24.43 & 25.76 & 25.39 \\
        Coverage Improvement (\%) & 58.96 & 44.84 & 48.84 \\
        Population Impact Score & 0.0283 & 0.0512 & 0.1144 \\
        Traffic Impact Score & 0.0142 & 0.0368 & 0.0457 \\
        Industrial Impact Score & 0.1025 & 0.0317 & 0.0735 \\
        Green Space Violations & 0 & 1 & 0 \\
        Spatial Entropy & 4.2485 & 4.2485 & 4.2485 \\
        \hline
    \end{tabular}
    \caption{Comparative Performance Analysis of Placement Strategies}
    \label{table:comparison}
\end{table}
\vspace{-60pt}
The PPO-based strategy outperforms the others in terms of population impact and traffic coverage while maintaining a high AQI improvement. The Greedy AQI approach achieves the highest overall AQI reduction but performs poorly in spatial distribution and long-term optimization. Random placement, while providing the largest coverage, does not effectively optimize air quality improvement in densely populated or high-traffic areas.
The training performance of the Proximal Policy Optimization (PPO) algorithm is visualized through various plots representing rewards, policy stability, loss values, and AQI improvements given in Figure~\ref{fig:comparison_analysis}. Each of these plots provides insights into different aspects of the learning process.

\vspace{-30pt}

\subsection{Multi-Dimensional Performance Comparison}

\begin{figure}[H]
    \centering
    \includegraphics[width=0.7\textwidth]{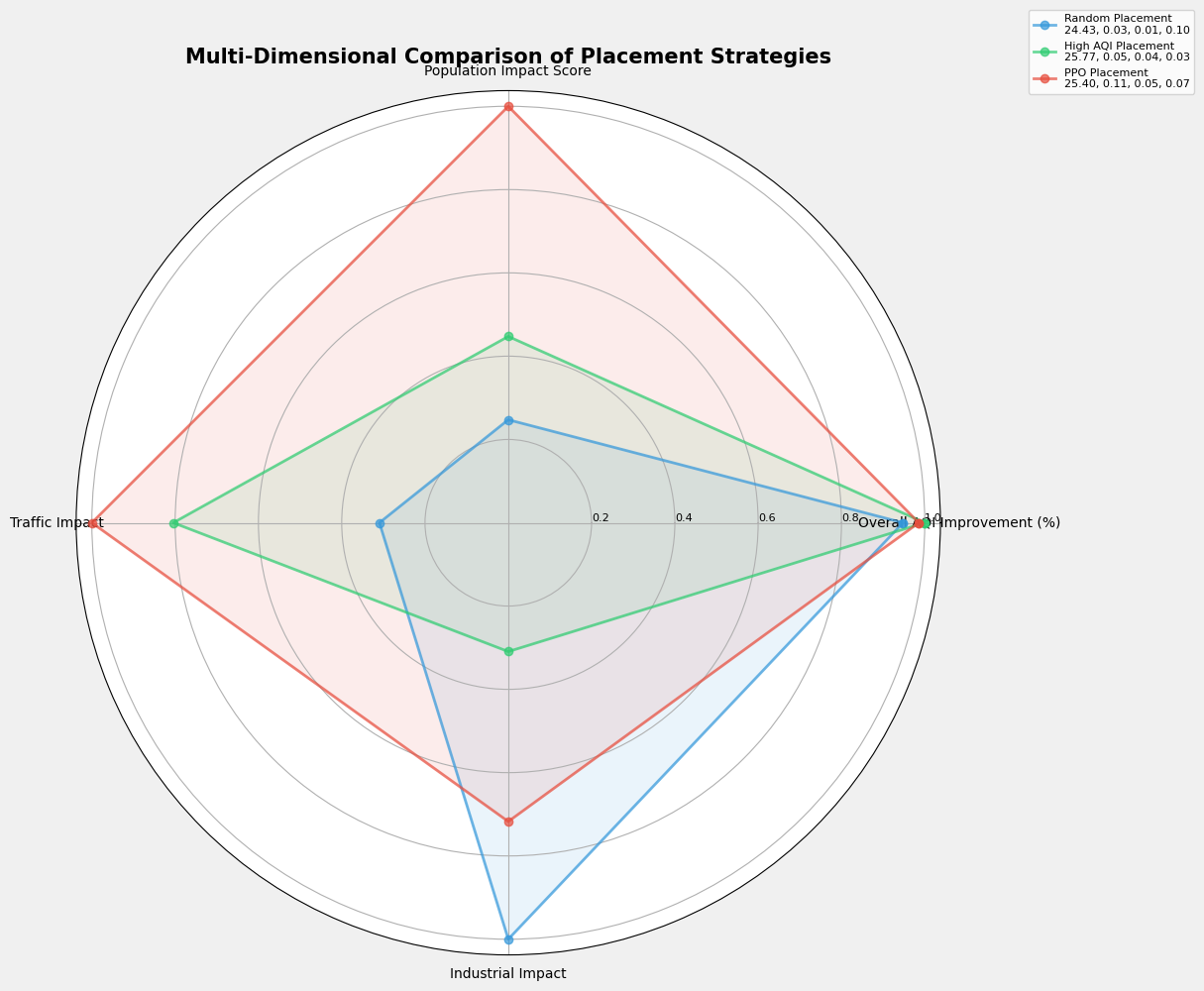}
    \caption{Multi-Dimensional Comparison of Placement Strategies}
    \label{fig:multidim_comparison}
\end{figure}

To further compare the strategies, we normalize key impact metrics and visualize them in a radar plot (Figure \ref{fig:multidim_comparison}). The normalization ensures all values lie within [0,1] for direct comparison.

\begin{equation}
    \text{Normalized Value} = \frac{X - X_{min}}{X_{max} - X_{min}}
\end{equation}
\vspace{-30pt}
\vspace{-60pt}
Figure~\ref{fig:multidim_comparison} presents a radar plot that compares the performance of the three booth placement strategies across multiple dimensions, including overall AQI improvement, population impact, traffic impact, and industrial impact. In this normalized plot, each axis represents one of the key metrics, and the area covered by each strategy’s plot reflects its overall effectiveness across these dimensions.

The PPO-based placement strategy exhibits the largest covered area in the radar plot, indicating that it achieves the highest performance in terms of population impact and traffic coverage. This suggests that the RL-based approach not only reduces AQI effectively but also strategically targets areas with high population density and heavy traffic, thus delivering a more balanced and socially beneficial outcome. In contrast, the greedy AQI method, while excelling in immediate AQI reduction, shows a smaller overall area on the radar plot. This reflects its tendency to focus on the most polluted areas without adequately considering other spatial factors, leading to a less balanced optimization. Meanwhile, the random placement strategy, although providing widespread spatial coverage, covers a relatively smaller area in high-impact dimensions, underscoring its inefficiency in targeting critical regions.

Overall, the multi-dimensional comparison plot underscores that the PPO-based strategy offers the best balance among competing objectives. It maximizes air quality improvement while ensuring that the benefits are well-distributed across population-dense and high-traffic areas, making it the most effective approach for urban air quality management.

\subsection{Discussion and Key Insights}

Our experimental analysis and comparative evaluation of booth placement strategies have yielded several critical insights, which we discuss in detail below. These insights not only validate the effectiveness of our PPO-based reinforcement learning approach but also highlight the limitations of heuristic methods and the importance of considering spatial distribution in urban air quality management.

One of the most significant observations is the \textbf{effectiveness of PPO-based optimization}. The PPO agent demonstrated a robust ability to balance multiple competing objectives, such as maximizing AQI improvement, enhancing population exposure to clean air, and ensuring efficient spatial distribution of the booths. Unlike heuristic methods that tend to target only the most polluted regions, the PPO-based strategy leverages a multi-objective reward function that takes into account factors such as population density, traffic, and industrial activity. This integrated approach is evident from the overall AQI improvement, which, although comparable to the greedy method, is achieved with a more balanced distribution that avoids clustering. The spatial entropy values indicate that the booths are neither too clustered nor too dispersed, suggesting that the PPO agent has learned to optimize placement in a way that addresses both localized and global air quality improvements. This balance is crucial in urban environments where pollution hotspots can shift rapidly due to dynamic factors, and our DRL approach provides the necessary adaptability.

In contrast, the \textbf{greedy high-AQI placement} method, while effective in yielding the highest immediate AQI improvement, presents significant trade-offs. This strategy prioritizes regions with the highest measured pollution levels, which, in a static sense, appears advantageous. However, the greedy approach neglects the long-term optimization and spatial balance required in a heterogeneous urban setting. By focusing solely on high-AQI cells, this method can lead to over-concentration of booths in a few areas, potentially missing opportunities to improve air quality in other critical regions. The resulting spatial imbalance may also lead to redundancy in the benefits obtained from closely clustered placements, reducing the overall efficiency of the intervention. Consequently, while the greedy method achieves impressive local improvements, it lacks the strategic foresight needed for sustainable urban planning.

The \textbf{random placement} strategy, although seemingly unsophisticated, provides valuable baseline insights. With booths distributed randomly across the city, this method ensures widespread coverage, which is beneficial for obtaining a rough approximation of the maximum possible spatial impact. However, its inherent lack of targeting means that many booths are placed in regions where their impact may be minimal—either due to low pollution levels or insufficient population density. The randomness results in significant variability in performance, as reflected in the lower population impact score and inconsistent AQI improvements. While random placement does ensure that no area is completely neglected, it ultimately falls short in terms of strategic effectiveness and fails to exploit data-driven insights that could lead to a more optimized distribution of resources.

Another critical insight emerges from our analysis of \textbf{spatial distribution considerations}. The PPO-based approach outperforms the heuristic methods by taking into account the spatial heterogeneity of urban environments. Unlike the greedy method, which concentrates on the most polluted cells, the PPO agent learns to identify and prioritize areas where booths can achieve the highest overall impact. This is particularly evident in the way the agent avoids placing booths in restricted regions, such as areas with high green space values, while focusing on high-traffic and high-population zones where air quality improvements can yield substantial public health benefits. The improved spatial coverage and balanced distribution lead to a more equitable dispersion of air quality improvements, thereby ensuring that the intervention benefits a larger portion of the urban population.

Additionally, our evaluation of the training metrics provides insights into the stability and convergence of the learning process. The reward curves, despite some fluctuations, show an upward trend, indicating that the agent is consistently learning a better placement strategy over time. The decreasing policy loss and entropy curves reflect that the PPO algorithm is successfully stabilizing its policy updates; the agent gradually transitions from exploratory behavior to a more deterministic policy, which suggests robust convergence. The observed fluctuations in the value loss, while indicative of ongoing adjustments in the value function, are typical in reinforcement learning and point to the inherent challenges of estimating long-term returns in a dynamic environment. Collectively, these trends confirm that our PPO agent is effectively balancing exploration and exploitation, adapting to complex urban conditions, and converging towards an optimal booth placement policy.

In summary, the comparative analysis clearly demonstrates that our DRL-based approach offers several advantages over traditional heuristic methods. The PPO-based strategy not only achieves competitive overall AQI improvement but also does so in a way that is spatially balanced and tailored to the dynamic nature of urban environments. The ability to integrate multiple environmental factors into the decision-making process is a key strength of our approach, making it highly relevant for real-world applications where pollution hotspots are both transient and spatially complex.

Moreover, our findings emphasize the importance of multi-objective optimization in urban planning. Effective air quality improvement requires more than just reducing pollution in isolated hotspots; it necessitates a comprehensive strategy that considers the broader urban context, including population distribution, traffic patterns, and industrial activity. By incorporating these factors into our reinforcement learning framework, we are able to deliver a more holistic and sustainable solution for urban air quality management.

Finally, while our results are promising, they also highlight areas for future improvement. The PPO agent's performance could potentially be enhanced by incorporating additional real-world factors such as wind speed and direction, which would provide a more dynamic and realistic representation of pollutant dispersion. Further research could also explore the scalability of our approach to larger urban areas and investigate hybrid models that combine RL with other optimization techniques to further refine booth placement strategies.

Overall, the insights gained from our analysis underscore the transformative potential of deep reinforcement learning in environmental management. Our work demonstrates that an intelligent, adaptive approach can significantly improve urban air quality by optimizing the placement of intervention infrastructure, thereby contributing to healthier, more sustainable urban living.

\section{Conclusion}

This study addresses the persistent challenge of urban air pollution by proposing a deep reinforcement learning (DRL)-based framework that optimizes the placement of air purifying booths. Urban areas are increasingly burdened by high levels of pollutants due to rapid urbanization, industrialization, and vehicular emissions. These factors have severe health, environmental, and economic impacts, particularly in densely populated megacities. Traditional air quality mitigation strategies often fall short in adapting to the dynamic, heterogeneous nature of urban environments. In contrast, our approach leverages the Proximal Policy Optimization (PPO) algorithm to learn an adaptive, data-driven strategy for booth placement that integrates multiple critical factors, including the Air Quality Index (AQI), population density, traffic flow, industrial activity, and green space constraints.

Our experimental evaluation demonstrates that the PPO-based method achieves an overall AQI improvement of approximately 25.40\% while maintaining superior spatial distribution compared to heuristic strategies such as random placement and greedy high-AQI placement. Although the greedy approach delivers the highest local AQI reduction, it tends to concentrate booth placements in limited areas, leading to uneven coverage and potential redundancies. Random placement, while ensuring broad spatial coverage, fails to prioritize high-impact zones, resulting in lower overall effectiveness. In contrast, the DRL-based strategy strikes a balanced trade-off by placing booths in a manner that not only reduces pollution but also maximizes population exposure to cleaner air and improves overall urban health outcomes. This multi-faceted optimization is further validated by improved population impact and traffic impact scores, as well as a well-balanced spatial entropy that suggests an efficient spread of booths across the city.

The proposed framework also contributes to the field by introducing a comprehensive evaluation paradigm that incorporates diverse metrics such as overall AQI improvement, spatial coverage, and multi-dimensional impact measures. By fusing heterogeneous data sources to create a high-resolution environmental grid, our approach provides a robust basis for real-time decision-making in complex urban settings. The adaptive nature of the reinforcement learning model enables it to respond dynamically to evolving urban conditions, ensuring that the placement strategy remains effective over time.

However, despite these promising results, our approach is not without limitations. The current model relies on a Gaussian influence function to approximate the impact of purification booths on AQI, which simplifies the intricate dynamics of pollutant dispersion. Critical factors such as wind speed and direction, local topography, and chemical reactions are not explicitly modeled, potentially affecting the precision of the booth impact estimation. Additionally, the effectiveness of our framework is heavily dependent on the quality and resolution of the input data, including AQI measurements and demographic information. Future work should focus on incorporating more sophisticated dispersion models and integrating meteorological data to better capture the real-world behavior of pollutants. Furthermore, extending our approach to larger urban areas and addressing scalability through hierarchical or multi-agent reinforcement learning techniques represent important directions for further research.

In summary, our research demonstrates the significant potential of deep reinforcement learning in optimizing urban air quality interventions. By employing the PPO algorithm within a multi-objective framework, we have developed an adaptive, efficient, and robust strategy for the strategic placement of air purifying booths. Our findings not only validate the superiority of the RL-based approach over traditional heuristic methods but also highlight the transformative role of data-driven decision-making in sustainable urban planning. Future advancements in incorporating dynamic environmental factors and enhancing computational efficiency will further solidify the application of AI-driven solutions in creating healthier, more resilient urban environments.

\section{Future Work and Limitations}  
Despite the effectiveness of our approach, certain limitations persist due to model simplifications, data constraints, and computational considerations. Addressing these limitations in future work could further enhance the robustness and applicability of our method.

\begin{itemize}
    \item \textbf{Model Simplifications:}  
    Our approach employs a Gaussian influence model to approximate the impact of purification booths on AQI levels. While this provides a mathematically convenient and interpretable framework, real-world pollution dispersion is influenced by complex factors, including chemical transformations, local topography, and dynamic meteorological conditions. Additionally, the booth effectiveness model assumes a uniform purification capacity, whereas real-world performance may vary based on external conditions such as humidity, temperature, and pollutant composition.  

    \textbf{Future Consideration:} Incorporating advanced pollution dispersion models, such as Computational Fluid Dynamics (CFD) simulations or data-driven spatial-temporal pollution forecasting models, could provide more precise estimations of booth impact and optimize placement strategies accordingly.

    \item \textbf{Lack of Meteorological Considerations (Wind Direction and Speed):}  
    A key limitation of our current model is the absence of meteorological factors, particularly wind direction and speed, which significantly influence pollution dispersion patterns. Without accounting for wind dynamics, the model assumes static AQI distributions, which may not accurately reflect real-world pollutant movement across urban areas.  

    \textbf{Future Consideration:} Incorporating wind data through meteorological simulations or real-time weather inputs could enable dynamic placement adjustments. Techniques such as Markov Decision Processes (MDPs) with external environmental state variables or physics-informed neural networks could be explored to adapt booth placements in response to changing wind conditions.

    \item \textbf{Policy and Deployment Constraints:}  
    Although our model optimizes booth placement based on spatial and environmental factors, practical deployment is subject to real-world constraints such as regulatory policies, public infrastructure limitations, and economic feasibility. Certain high-impact locations identified by the model may not be feasible due to land-use restrictions or high installation costs.  

    \textbf{Future Consideration:} Future studies could integrate economic and policy-aware optimization frameworks, incorporating budget constraints and regulatory considerations into the RL training process. Game-theoretic approaches or participatory urban planning models could also be explored to ensure practical feasibility and stakeholder alignment.

\end{itemize}

While these limitations present opportunities for further refinement, our results demonstrate the effectiveness of DRL-based optimization for intelligent urban air quality interventions. By addressing these challenges, future iterations of this framework could provide even more precise, adaptable, and scalable solutions for pollution mitigation in complex urban environments.

\section*{Funding}
This research did not receive any specific grant from funding agencies in the public, commercial, or not-for-profit sectors.

\section*{Ethical Statement}
This study does not involve human participants or animals, and thus no ethical approval is required.

\section*{CRediT authorship contribution statement}

\section*{Declaration of Competing Interest}
No conflict of interest.

\section*{Data Availability}
Data will be made available on request.

\section*{References}
\begin{enumerate}
   \setlength{\itemindent}{-1em}
   \raggedright
   \item Gaybullaev, O. (2024). An analysis of the economic impacts of air pollution in the urban areas: bibliographic review. Economica, 15(3–4), 53–68. https://doi.org/10.47282/economica/2024/15/3-4/15028
   \item Mahala, K. R. (2024). The impact of air pollution on living things and Environment: A review of the current evidence. World Journal Of Advanced Research and Reviews, 24(3), 3207–3217. https://doi.org/10.30574/wjarr.2024.24.3.3929
   \item Nawaz, H., Muhammad Umar, M. U., Parvaiz, F., Shehzad, M. T., \& Afraz, I. (2024). Impacts of Climatic Changes and Air Pollution on Public health and Environment. 3(1). https://doi.org/10.37939/jhcc.v3i1.12
   \item Wasi, T. A. (2024). Long-term Effects of Air Pollution on Respiratory Health in Urban Populations. 1(6), 14–25. https://doi.org/10.70008/nhj.v1i06.31
   \item Kumar, P., Garg, A., Sharma, K., Nadeem, U., Sarma, K., Gupta, N. C., Kumar, A., \& Pandey, A. K. (2024). Seasonal and Spatial Variations in Particulate Matter, Black Carbon and Metals in Delhi, India’s Megacity. Urban Science. https://doi.org/10.3390/urbansci8030101
   \item Singh, B., Gahlaut, V., \& Rani, N. (2024). Seasonal and annual emission trends of PM2.5 and PM10 over the national capital Delhi from 2015 to 2020. Journal of Air Pollution and Health. 
   \item Kumar, V. (2024). Analyzing Air Pollution in Urban Delhi: A Quantitative Assessment of Particulate Matter (PM) Concentrations. International Journal For Multidisciplinary Research. https://doi.org/10.36948/ijfmr.2024.v06i02.16872	
   \item Weng, Z., Zhang, J., \& Zhao, X. (2024). The Strategies of Air Quality Improvement in Urban Governance: A Case Study of Beijing. Journal of Education, Humanities and Social Sciences, 33, 36–41. https://doi.org/10.54097/fqg1p983
   \item Gebreyesus, T., Borgemeister, C., Herrero‐Jáuregui, C., \& Kelboro, G. (2024). Transforming Urban Air Quality: Green Infrastructure Strategies for the Urban Centers of Ethiopia. Environmental Pollution, 363, 125244. https://doi.org/10.1016/j.envpol.2024.125244
   \item Sah, S. K. (2024). Public health policy responses to air pollution in india: challenges and strategies for sustainable development. Visioner, 16(2), 163–170. https://doi.org/10.54783/jv.v16i2.1057
   \item Liao, J., \& Kim, H.-Y. (2024). Analyzing the Relationship between Green Infrastructure and Air Quality Issues—South Korean Cases. Land, 13(8), 1263. https://doi.org/10.3390/land13081263
   \item Akomolafe, O. O., Olorunsogo, T., Anyanwu, E. C., Osasona, F., Ogugua, J. O., \& Daraojimba, O. H. (2024). Air quality and public health: a review of urban pollution sources and mitigation measures. https://doi.org/10.51594/estj.v5i2.751
   \item Hackenberger, B. K., Djerdj, T., \& Hackenberger, D. K. (2025). Advancing Environmental Monitoring through AI: Applications of R and Python. https://doi.org/10.5772/intechopen.1007683
   \item Milutinović, M. (2024). Machine learning in environmental monitoring. Facta Universitatis, 155. https://doi.org/10.22190/fuwlep241029014m
   \item Banasode, P., Pawar, M., Patki, V., R, S., Patange, S., \& Sangolli, S. (2024). Data-Driven Waste Management: Machine Learning and Excess Reprocessing for Optimized Environmental Sustainability. 1–5. https://doi.org/10.1109/innova63080.2024.10847036
   \item Adekoya, M., \& Ogbolumani, O. A. (2025). Intelligent Waste Management Optimization Through Machine Learning Analytics. 2(1), 7–26. https://doi.org/10.70882/josrar.2025.v2i1.25
   \item Ramya Shree, A. N., \& Reddy, C. S. B. (2024). Air Quality Management in Smart Cities by leveraging Machine Learning Techniques. 978–983. https://doi.org/10.1109/icacrs62842.2024.10841590
   \item Zhao, C. (2024). Application of Reinforcement Learning in Complex Environmental Decision-making Problems. Journal of Computing and Electronic Information Management, 15(3), 104–108. https://doi.org/10.54097/a1nnjq52
   \item Barrionuevo, A. M., Yanes Luis, S., Gutiérrez Reina, D., \& Toral, S. L. (2024). Optimizing Plastic Waste Collection in Water Bodies Using Heterogeneous Autonomous Surface Vehicles with Deep Reinforcement Learning. https://doi.org/10.48550/arxiv.2412.02316
   \item Siddiki, A., \& Arif, I. (2024). AI-Driven Adaptive Ventilation Systems For Real-Time Pollution Control In Industrial And Urban Settings: A Systematic Review. 1(01), 56–73. https://doi.org/10.70008/jeser.v1i01.48
   \item Abbass, M., Akhai, S., Abbas, U., Jafri, R., \& Arif, S. Mohd. (2024). AI-Enabled Sustainable Urban Planning and Management. Advances in Computational Intelligence and Robotics Book Series, 233–260. https://doi.org/10.4018/979-8-3693-4252-7.ch012
   \item Palma-Borda, J., Guzmán, E., \& Belmonte, M. (2025). Cooperative Patrol Routing: Optimizing Urban Crime Surveillance through Multi-Agent Reinforcement Learning. https://doi.org/10.48550/arxiv.2501.08020
   \item Taha, H., \& Abdelhadi, A. (2025). HEPPO: Hardware-Efficient Proximal Policy Optimization -- A Universal Pipelined Architecture for Generalized Advantage Estimation. https://doi.org/10.48550/arxiv.2501.12703
   \item Rashmita, M., Chilukuri, K. C., Kumar, R., Idhyavathi, T. V., Suneetha, R., Prasad, V. L. M., \& Samatha, B. (2024). Proximal Policy Optimization for Efficient Channel Allocation with Quality of Service (QoS) in Cognitive Radio Networks. International Journal of Experimental Research and Review, 46, 326–341. https://doi.org/10.52756/ijerr.2024.v46.026
   \item Jia, L., Su, B., Xu, D., Wang, Y., Fang, J., \& Wang, J. (2024). Policy Optimization Algorithm with Activation Likelihood-Ratio for Multi-agent Reinforcement Learning. Neural Processing Letters, 56(6). https://doi.org/10.1007/s11063-024-11705-x

   \item Hura, V., \& Monastyrskii, L. S. (2024). Influence of Air Quality Model Parameters on Pollution Concentration. Štučnij Ìntelekt, 29(AI.2024.29(4)), 207–217. https://doi.org/10.15407/jai2024.04.207
   \item Prataviera, E., \& De Carli, M. (2024). A review on outdoor urban environment modelling. Journal of Physics, 2893(1), 012024. https://doi.org/10.1088/1742-6596/2893/1/012024
   \item Sihorwala, Z. (2024). AQINet: A multimodal deep convolutional neural network to predict Air Quality Index via satellite imagery and meteorological data. 1, 46–50. https://doi.org/10.62329/gkjr4572
   \item Vijayalakshmi, A., Abishek.B, E., Rubi, J., Dhivya, J., Kavidoss, K., \& Ram A.S, A. (2024). Machine Learning-Based Prediction and Analysis of Air and Noise Pollution in Urban Environments. 1080–1085. https://doi.org/10.1109/icscss60660.2024.10625644
   \item Mohammed, A., Saif, O., Abo-Adma, M. A., \& Elazab, R. (2024). Multiobjective optimization for sizing and placing electric vehicle charging stations considering comprehensive uncertainties. Energy Informatics, 7(1). https://doi.org/10.1186/s42162-024-00428-x

\noindent 
\end{enumerate}

\end{document}